\documentclass[11pt]{article}

\usepackage[margin=1in]{geometry}
\usepackage{times}

\usepackage{amsmath,amsfonts,bm,amsthm}

\def\1{\bm{1}}

\DeclareMathAlphabet{\mathsfit}{\encodingdefault}{\sfdefault}{m}{sl}
\SetMathAlphabet{\mathsfit}{bold}{\encodingdefault}{\sfdefault}{bx}{n}

\usepackage[round,authoryear]{natbib}
\setcitestyle{aysep={,}}
\renewcommand{\cite}{\citep}  

\usepackage{hyperref}
\hypersetup{
  colorlinks=true,
  linkcolor=blue!60!black,
  citecolor=blue!60!black,
  urlcolor=blue!60!black
}
\usepackage{url}
\usepackage{algorithm}
\usepackage{algpseudocode}
\algrenewcommand\algorithmiccomment[1]{\hfill\small\textcolor{gray}{\textit{// #1}}}
\usepackage{booktabs}
\usepackage{multirow}
\usepackage{graphicx}
\usepackage{colortbl,xcolor}
\usepackage{soul}
\colorlet{llgray}{lightgray!40}
\sethlcolor{llgray}
\usepackage{mathtools}
\usepackage{amsmath}
\usepackage{amssymb}
\usepackage{amsthm}
\usepackage{caption}
\usepackage{subcaption}
\usepackage{wrapfig}
\usepackage{placeins}
\usepackage{float}
\usepackage{siunitx}
\usepackage{tabularx}
\usepackage{bbm}
\usepackage{authblk}
\usepackage{enumitem}

\usepackage{xcolor}

\definecolor{ourscolor}{HTML}{E6F2FF}

\theoremstyle{plain}

\theoremstyle{definition}

\theoremstyle{remark}

\title{\bfseries Stream4D: 4D-Consistency for \\ Streaming Autoregressive Diffusion Video Models}
\author[1]{Yuanhao Ban}
\author[2]{Jiaqi Feng}
\author[1]{Hengguang Zhou}
\author[1]{Xiaohuan Pei}
\author[1]{Justin Cui}
\author[1]{Cho-Jui Hsieh}

\affil[1]{UCLA}
\affil[2]{Tsinghua University}

\begin{document}

\maketitle


\begin{abstract}
Streaming autoregressive diffusion models enable real-time, long-horizon video generation, but their training objectives optimize local frame prediction rather than the geometry and dynamics of a coherent world: long rollouts accumulate geometric drift and degrade into static or unnatural motion. Recent bidirectional approaches address this problem using rewards signals built upon 3D Gaussian-Splatting reconstruction. However, a single rigid 3d reconstruction cannot model a dynamic scene, so this critic \emph{penalizes} genuine object motion as reconstruction error and is maximized by freezing the video. This shortcut is especially detrimental in the AR setting, where each chunk can propagate an already-static configuration.
In this work, we propose Stream4D, which replaces the static critic with a feed-forward 4D reconstruction reward that explicitly models scene dynamics, allowing coherent motion to receive high consistency rewards. To further guide motion magnitude and quality, we add a motion prior that rewards natural scene-flow magnitude while penalizing jitter and non-rigid artifacts. Our final recipe combines these two terms with a lightweight perceptual anchor. Across various  autoregressive video backbones and various generation horizons, Stream4D improves 4D reconstruction quality, preserves motion more effectively, and achieves higher human-aligned preference.  Project page: https://banyuanhao.github.io/Stream4D/
\end{abstract}

\section{Introduction}
\label{sec:intro}

\begin{figure*}[t]
\centering
\includegraphics[width=\textwidth]{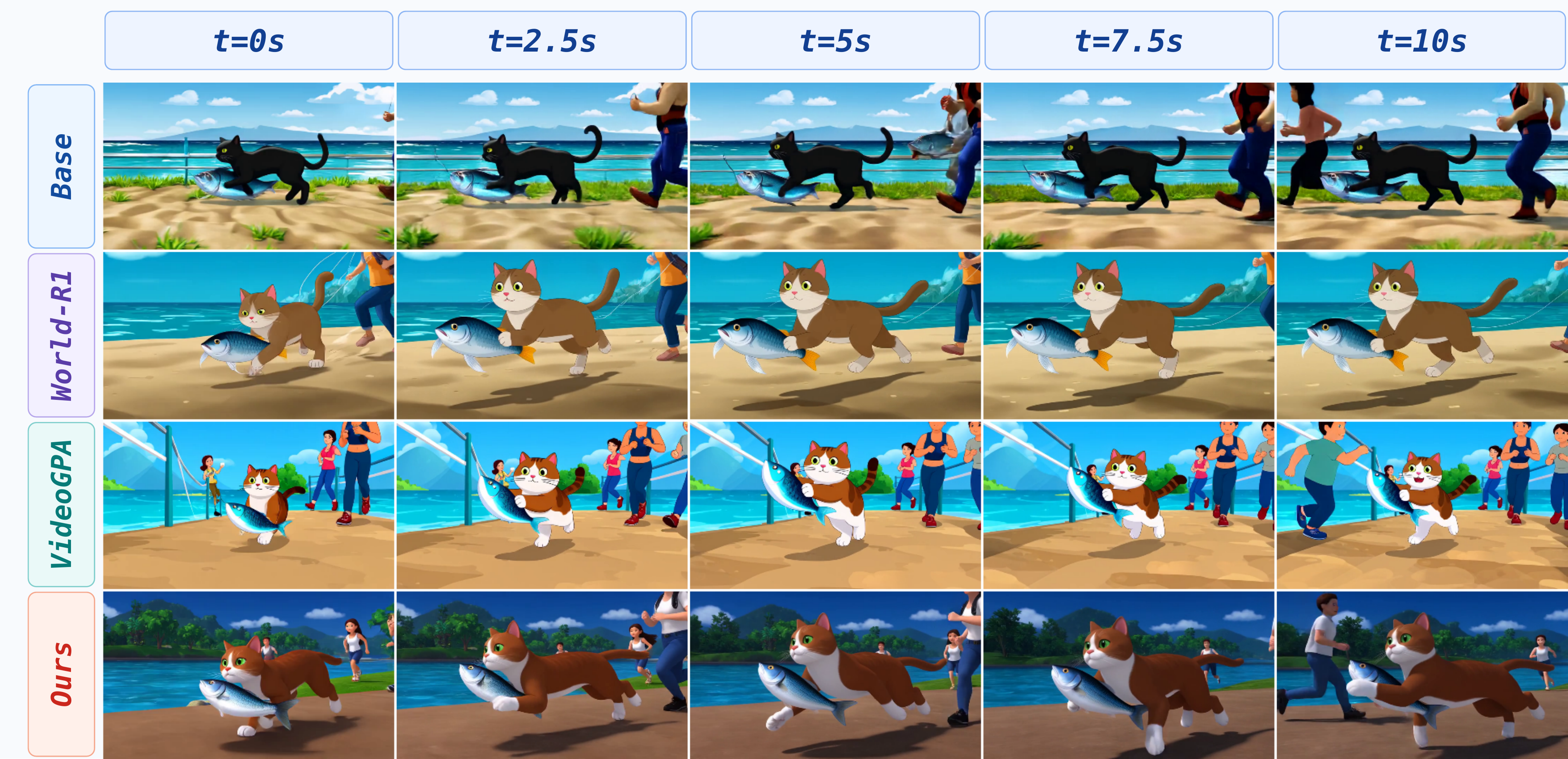}
\caption{\textbf{Static-3D rewards freeze the scene; a 4D reward keeps it moving.} Five uniformly-spaced frames from a $10.3$\,s LongLive rollout (``a cat running away with a fish while people chase behind''). The distilled base contains large motion but the cat and fish drift across frames; the static-3DGS rewards World-R1 and VideoGPA reduce the scene to a more rigid, low-motion configuration. Stream4D keeps the cat running with clear forward motion while preserving a coherent subject and scene.}
\label{fig:teaser}
\end{figure*}

Autoregressive (AR) diffusion video models~\cite{huang2025self, yang2025longlive, cui2025self, zhu2026causal, yesiltepe2025infinityrope} generate videos chunk by chunk, enabling extendable temporal horizons. Combined with recent few-step backbones, such as Wan2.1~\cite{wan2025wan}, they offer a practical path toward real-time streaming video generation for embodied agents, simulators, and interactive applications. However, existing objectives do not sufficiently enforce consistent world dynamics over long rollouts. A successful rollout should maintain both \textbf{4D consistency} geometry, depth, object identity, and camera motion, and a \textbf{natural motion prior} with realistic magnitude and temporal smoothness. In practice,  AR models often accumulate errors over time, causing scale drift, inconsistent depth relationships, and implausible scene or camera motion.

Recent reinforcement-learning approaches, including VideoGPA~\cite{du2026videogpa} and World-R1~\cite{wang2026world}, attempt to improve geometric consistency using rewards derived from static 3D Gaussian Splatting (3D-GS) reconstruction. However, the rigid 3d reconstruction model prior cannot represent a dynamic scene. Any genuine object motion appears as reconstruction error and is punished by the reward, which results in frozen scene and permitting only camera movement, as illustrated in Figure~\ref{fig:teaser}. Moreover, these rewards were originally developed for bidirectional video-generation settings, whereas their limitations become more pronounced in autoregressive generation. An AR model can attend only to previously generated frames and therefore has less temporal context with which to maintain coherent scene dynamics. Under this restricted information, collapsing to a static scene becomes an even easier shortcut: once motion is suppressed in early frames, subsequent chunks can continue propagating the same rigid configuration while still receiving a high reconstruction-based reward.

To address the issues mentioned above, we introduce \textbf{Stream4D}, a reinforcement-learning framework that distills 4D-consistency and natural-motion priors into streaming AR video models. Our key insight is that recent feed-forward 4D Gaussian Splatting (4D-GS) models, such as MoVieS~\cite{lin2025movies}, provide a useful learned prior over how dynamic scenes evolve in space and time. Given a rollout sampled from the student policy, we reconstruct it using a pretrained 4D-GS model, re-render the resulting dynamic scene from its estimated camera trajectory, and measure perceptual agreement between the rendered frames and the original rollout. This reconstruction score rewards rollouts that can be explained by coherent geometry, motion, and camera dynamics, serving as a good 4d-consistency score.

To further imporve the motion quality and guide the tensity, Stream4D adds a  Gaussian motion reward peaking at the natural motion tensity that penalizes both insufficient and excessive motion. We further modulate it using temporal-smoothness and rigidity-quality factors, preventing the policy from reaching the target motion magnitude through jitter, blur, or geometrically implausible shortcuts. Finally, we include a lightweight image reward model as a perceptual anchor to preserve the visual fidelity of the original model. The three reward components are standardized independently using $z$-normalization and then combined additively, producing a lightweight recipe that transfers across multiple distilled AR backbones. We evaluate Stream4D on three streaming AR architectures, covering both 5s and 10s video generation. Across Self-Forcing~\cite{huang2025self}, Causal-Forcing~\cite{zhu2026causal}, and LongLive~\cite{yang2025longlive} backbones, Stream4D improves 4D-PSNR from 16.88, 15,44 and 17.44 to 20.34, 20.97 24.20, respectively, while preserving motion and maintaining strong performance under general-purpose video reward models. 

Our contributions are threefold:
\begin{itemize}
\item We identify accumulated 4D inconsistency as a central failure mode of distilled streaming AR video models and formulate its mitigation as a reinforcement-learning problem.

\item We introduce Stream4D, a transferable training objective that combines structured 4D reconstruction consistency, a target-centered natural-motion prior, and perceptual anchoring.

\item We demonstrate consistent improvements across three distilled AR backbones, including gains of up to \textbf{6.76\,dB} in 4D-PSNR, while preserving motion and general video quality over both short and long generation horizons.

\end{itemize}

\section{Related Work}
\label{sec:related}

\subsection{Autoregressive Diffusion Video Models}
Diffusion-based video generation has progressed from joint-frame DiTs~\cite{yang2024cogvideox, bao2024vidu, kong2024hunyuanvideo, wan2025wan} toward streaming autoregressive variants that emit clips chunk-by-chunk. Early teacher-forced AR approaches~\cite{hu2024acdit, gao2024ca2} suffered from error accumulation; Diffusion Forcing~\cite{chen2024diffusion}, CausVid~\cite{yin2025slow} with distribution-matching distillation~\cite{yin2024one}, Self-Forcing~\cite{huang2025self}, LongLive~\cite{yang2025longlive}, Infinity-RoPE~\cite{yesiltepe2025infinityrope}, and Causal-Forcing~\cite{zhu2026causal} progressively closed the train--test gap via per-frame noise schedules, block-causal attention, self-rollout, and AR teachers. The resulting distilled streaming models~\cite{sun2025worldplay} support real-time generation and double as video world models, but their supervision provides no incentive for the 3D scene to remain coherent across the rollout explicitly.

\subsection{Reinforcement Learning for Image and Video Generation}
Dance-GRPO~\cite{xue2025dancegrpo} and Flow-GRPO~\cite{liu2025flowgrpo} run on-policy GRPO over reverse-process trajectories, requiring log-probability estimation along the sampling chain and full trajectory storage. DiffusionNFT~\cite{zheng2025diffusionnft} reformulates RL on the \emph{forward} process via negative-aware fine-tuning; WorldCompass~\cite{wang2026worldcompass} adapts it to autoregressive world models, and Astrolabe~\cite{astrolabe_2026} brings it to distilled streaming AR backbones with a rolling KV cache and a multi-reward objective.

\subsection{Geometry-Aware Rewards for Video Generation}
A complementary line uses 3D reconstruction as a critic. World-R1~\cite{wang2026world} combines a feed-forward depth/camera estimator~\cite{lin2025depth}, a per-clip 3D Gaussian-Splat reconstruction~\cite{ye2025gsplat}, and a VLM critic~\cite{Qwen3-VL} into a 3D-aware reward for \emph{bidirectional} text-to-video under Flow-GRPO; VideoGPA~\cite{du2026videogpa} extends this direction with preference alignment and a static-3D evaluation suite~\cite{kupyn2025epipolar, liu2025improvingvideogenerationhuman}. Both reconstruct the scene as a \emph{single rigid} 3D Gaussian-Splat, which cannot represent a dynamic scene; consequently inter-frame object motion is structurally \emph{penalized} as 3D inconsistency, and the reward is maximized by suppressing motion rather than merely being indifferent to it. We recast this \emph{3D-consistency} framing as \emph{4D-consistency} by replacing the reconstructor with MoVieS~\cite{lin2025movies}, a feed-forward 4D-GS model that factors the scene into a canonical point cloud plus per-frame attribute overrides and scene-flow offsets.

\section{Method}
\label{sec:method}

Stream4D is a reinforcement-learning recipe for distilled autoregressive (AR) video models. Each rollout is graded against a feed-forward 4D Gaussian-Splatting reconstruction of itself, through 4D reconstruction consistency, a gated motion term, and a lightweight perceptual anchor. Then the rewards are summed under per-axis $z$-normalization and optimized with the forward-process DiffusionNFT loss.

\subsection{Preliminaries}

\label{sec:method:prelims}

\paragraph{Forward-process streaming-AR RL.} We build on Astrolabe~\cite{astrolabe_2026}, which combines rolling-KV-cache generation with the DiffusionNFT~\cite{zheng2025diffusionnft} update for distilled AR video models. From a shared context, we sample a group of $G$ candidate rollouts $\{W^{(i)}\}_{i=1}^{G}$ for group-wise reward normalization. For each rollout and denoising step, let $v_{\theta}$ and $v_{\theta_{\mathrm{old}}}$ denote the current and reference-model velocity predictions in the latent-video space. We write
$v_{\theta}, v_{\theta_{\mathrm{old}}}, v^{+}, v^{-}, v_{\mathrm{target}} \in \mathbb{R}^{d}$ after flattening the latent tensor.
DiffusionNFT forms positive and negative interpolated velocity predictions
\begin{equation}
v^{+}
=
(1-\beta)v_{\theta_{\mathrm{old}}}
+
\beta v_{\theta},
\qquad
v^{-}
=
(1+\beta)v_{\theta_{\mathrm{old}}}
-
\beta v_{\theta},
\label{eq:nft_interp}
\end{equation}
and optimizes
\begin{equation}
\mathcal{L}_{\mathrm{policy}}
=
\tilde{r}\,
\left\lVert v^{+}-v_{\mathrm{target}}\right\rVert_2^2
+
(1-\tilde{r})\,
\left\lVert v^{-}-v_{\mathrm{target}}\right\rVert_2^2,
\label{eq:nft_policy_loss}
\end{equation}
where $v_{\mathrm{target}}$ is the diffusion velocity target and $\tilde{r}\in[0,1]$ is the normalized rollout reward. Training details are provided in Sec.~\ref{sec:exp:setup}.

\paragraph{MoVieS 4D-GS reconstruction.} Given a candidate rollout $W^{(i)}=\{W^{(i)}_t\}_{t=1}^{T}$, we first estimate per-frame cameras with StreamVGGT~\cite{streamvggt_2026}. We then run MoVieS~\cite{lin2025movies}, a feed-forward 4D Gaussian-Splatting reconstructor, conditioned on the sampled frames and estimated cameras. MoVieS represents the video as a dynamic Gaussian scene: a canonical set of 3D Gaussians together with time-dependent deformation and appearance parameters that allow the scene to move and change across frames. In our pipeline, we use two outputs from this reconstruction. First, MoVieS re-renders reconstructed frames $\tilde W^{(i)}_t$ from the estimated camera at each time $t$, which we use for the reconstruction reward. Second, MoVieS provides a per-pixel 3D motion field $P \in \mathbb{R}^{T \times H \times W \times 3}$ and confidence map $\mathrm{conf}\in[0,1]^{T\times H\times W}$, which we use to compute motion magnitude, smoothness, and rigidity.

\subsection{Stream4D Reward design}
\label{sec:method:reward}

This section first defines the three reward components and then explains how they are combined into the final advantage.

\paragraph{4D-GS reconstruction $R_{\text{recon}}$.}

For each candidate rollout, we subsample 26 frames and use the MoVieS reconstruction described above to render the corresponding reconstructed frames from the estimated cameras. We compare each rendered frame with the original generated frame and define the reconstruction reward as the clipped average perceptual agreement:
\begin{equation}
R_{\text{recon}} =
\mathrm{clip}\!\left(
1 - \frac{1}{T}\sum_t \mathrm{LPIPS}(\tilde W^{(i)}_t, W^{(i)}_t),
0, 1
\right).
\label{eq:r_recon}
\end{equation}
Here $W^{(i)}_t$ is the $t$-th frame of candidate rollout $i$, $\tilde W^{(i)}_t$ is the corresponding frame re-rendered from the MoVieS 4D-GS reconstruction using the estimated camera, and LPIPS~\cite{zhang2018unreasonable} denotes the Learned Perceptual Image Patch Similarity distance. LPIPS measures perceptual discrepancy in a pretrained deep feature space, making it less sensitive than pixel-wise losses to small low-level misalignments while still penalizing visual reconstruction errors. Thus, the reward is high when the rollout can be explained by a coherent 4D reconstruction and low when the video contains inconsistent geometry, drifting object identity, or motion that cannot be organized into a stable dynamic scene.

\begin{wrapfigure}{r}{0.42\textwidth}
\vspace{-4mm}
\centering
\includegraphics[width=0.4\textwidth]{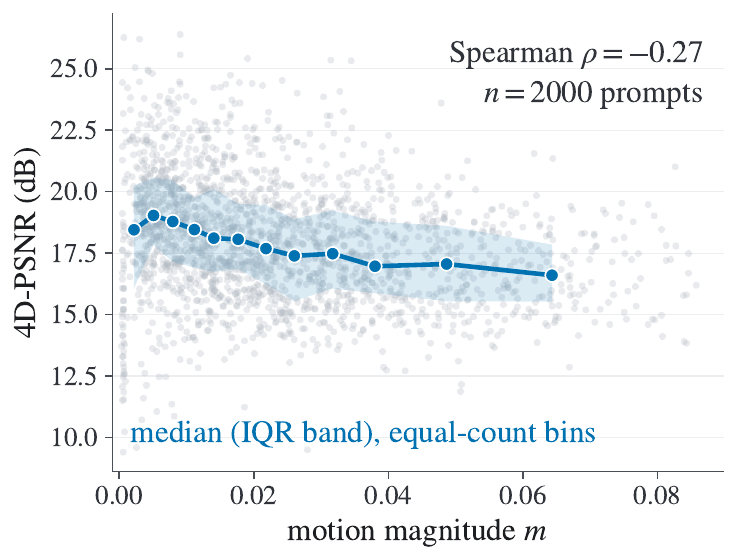}
\caption{Unlike a static 3D reward, which punishes motion outright, our dynamic 4D-GS backbone does not penalize motion. Per-prompt 4D-PSNR only drifts weakly downward with motion magnitude $m$.}
\label{fig:mechanism}
\vspace{-4mm}
\end{wrapfigure}

\paragraph{Gated motion conjunction $R_{\text{mot}}$.}
Unlike the static 3D reward, the 4D reconstructor does not punish motion. However, lower-motion clips are still slightly easier to reconstruct. As shown in Fig.~\ref{fig:mechanism}, there is a weak Spearman $\rho=-0.27$ between per-prompt 4D-PSNR and the motion-gate input $m$ computed on the Self-Forcing base model over the held-out prompts. To address this, we add an explicit, quality-weighted motion reward that guides both motion intensity and motion quality:
\begin{equation}
    R_{\text{mot}} \;=\; g(m)\cdot \mathrm{smooth}\cdot \mathrm{rigid},
    \label{eq:r_mot}
\end{equation}
where $m$ is the motion intensity, $g(\cdot)$ is a Gaussian gate, and $\mathrm{smooth}$ and $\mathrm{rigid}$ are quality factors that penalize jitter and spatially erratic flow. We compute these terms from the MoVieS motion field $P$ and confidence map $\mathrm{conf}$ defined in Sec.~\ref{sec:method:prelims}. Let $v_t=P_{t+1}-P_t$ be the per-pixel 3D scene-flow velocity and $\|v_t\|$ its magnitude. The dynamic mask $\mathcal{D}$ contains the top $20\%$ fastest pixel-time entries in the clip. We write
$\mathrm{mean}_{\mathcal{D}}^{\mathrm{conf}}[f]
= \bigl(\sum_{(p,t)\in\mathcal{D}}\mathrm{conf}_{p,t}f_{p,t}\bigr)/
\bigl(\sum_{(p,t)\in\mathcal{D}}\mathrm{conf}_{p,t}\bigr)$
for the confidence-weighted mask mean and $\mathrm{mean}_{\mathcal{D}}[f]$ for the uniform one. The clip-level motion magnitude is
$m=\mathrm{mean}_{\mathcal{D}}^{\mathrm{conf}}\|v_t\|$.

\textit{The motion gate} is a peaked Gaussian penalizing both static collapse and runaway blur:
\begin{equation}
    g(m) \;=\; \exp\!\left(-\tfrac{(m-m_{\text{nat}})^2}{2\sigma^2}\right),
    \label{eq:gauss_gate}
\end{equation}
where the target $m_{\text{nat}}$ is the median per-clip $m$ of base rollouts. We choose the Gaussian kernel because it penalizes both overly aggressive and nearly static motion, thereby encouraging motion intensity to remain within a natural range. We also evaluate alternative formulations, such as a linear motion function, in the ablation study in Sec.~\ref{sec:exp}.


\textit{Smoothness} penalizes excessive aggregate motion and deformation and is formulated as follows:

\begin{equation}
    \mathrm{smooth} \;=\; \exp\!\left(-\,\frac{\mathrm{mean}_{\mathcal{D}}^{\mathrm{conf}}\,\|v_{t+1}-v_t\|}{m+c}\right).
    \label{eq:smooth}
\end{equation}
where $c=10^{-4}$ is a small constant for numerical stability.

\textit{Rigidity} penalizes spatially erratic scene flow. Although real objects can deform or articulate, nearby points on the same moving object usually have locally coherent 3D velocities. We therefore compute the spatial finite differences of the per-pixel scene-flow velocity $v_t$ within the dynamic mask $\mathcal{D}$. Large $\|\nabla_x v_t\|$ or $\|\nabla_y v_t\|$ indicates neighboring pixels moving with sharply different 3D velocities, which often corresponds to tearing, jitter, or melting artifacts. We map this roughness to a bounded quality factor:
\begin{equation}
    \mathrm{rigid} \;=\; \exp\!\left(-\,\tfrac{k_{\text{rough}}}{2}\,\bigl(\mathrm{mean}_{\mathcal{D}}\|\nabla_x v_t\| \,+\, \mathrm{mean}_{\mathcal{D}}\|\nabla_y v_t\|\bigr)\right), \quad k_{\text{rough}}\!=\!400.
    \label{eq:rigid}
\end{equation} 
This encourages locally coherent motion while still allowing smooth non-rigid or articulated dynamics.

\paragraph{Perceptual anchor $R_{\text{hpsv2}}$.}
We use HPSv2~\cite{wu2023hpsv2} to score the aesthetics of the generated frames directly. The reconstruction and motion terms constrain geometry and dynamics but leave per-frame appearance unconstrained. The perceptual anchor addresses this by tying the reward to human aesthetic preference, keeping generations close to the visual fidelity of the base model. 
\paragraph{Z-norm reward ensembling.}
Following~\cite{liu2026gdpo}, our final reward sums three $z$-normalized axes evaluated on each candidate rollout $W^{(i)}$:
\begin{equation}
    R(W^{(i)}) \;=\; w_{\text{recon}}\,\tilde z\bigl[R_{\text{recon}}\bigr] \;+\; w_{\text{mot}}\,\tilde z\bigl[R_{\text{mot}}\bigr] \;+\; w_{\text{hpsv2}}\,\tilde z\bigl[R_{\text{hpsv2}}\bigr],
    \label{eq:reward}
\end{equation}
where $\tilde z[\cdot]$ is the per-axis $z$-score over group. Following the DiffusionNFT~\cite{zheng2025diffusionnft}, we compute the group-centered advantage $A^{(i)} = R(W^{(i)}) - \tfrac{1}{G}\sum_j R(W^{(j)})$ with clipped affine normalization,
\begin{equation}
    \tilde r^{(i)} \;=\; \mathrm{clip}\!\left(A^{(i)}/A_{\max},\,-1,\,1\right)/2 + 1/2,
    \label{eq:r_tilde}
\end{equation}
where $A_{\max}\!=\!5$ is a fixed constant.
\section{Experiments}
\label{sec:exp}

\subsection{Setup}
\label{sec:exp:setup}
\noindent\textbf{Backbones and training.} We evaluate Stream4D on three distilled autoregressive video backbones: Self-Forcing~\cite{huang2025self}, Causal-Forcing~\cite{zhu2026causal}, and LongLive~\cite{yang2025longlive}, using the Stream4D reward formulation in Eq.~\ref{eq:reward}. Self-Forcing and Causal-Forcing use their native 81-frame setting, corresponding to approximately 5s at 16fps, while LongLive uses a 10.3,s window of 165 frames. A LoRA adapter is trained over the frozen base with the forward-process NFT loss of Sec.~\ref{sec:method:prelims}. We trained every method for 150 epochs and reported the results on final checkpoints. We randomly sampled training prompts from VidProM~\cite{wang2024vidprom}. 
Full hyperparameters, the reward-stack implementation, and compute details are in Appendices~\ref{sec:supp_reward} and~\ref{sec:supp_hyper}; the full procedure is Algorithm~\ref{alg:streaming_diffusionnft}.

\noindent\textbf{Evaluation protocol.}
To stress-test the motion-preservation capabilities of our method, we construct a 500-prompt \emph{motion-prominent} subset of VidProM using a motion filter and a keyword-based filter targeting multi-agent interactions and object motion. We also report results on 500 randomly selected prompts in Appendix~\ref{sec:supp_rand500}. We evaluate with three groups of metrics:
\begin{itemize}[leftmargin=1.2em,itemsep=1pt,topsep=1pt]
  \item \emph{MoVieS Recon} (PSNR\,$\uparrow$ / SSIM\,$\uparrow$ / LPIPS\,$\downarrow$): re-render error of the MoVieS 4D-GS reconstruction. Full computation pipeline and the definition of PSNR, SSIM and LPIPS could be found in Appendix~\ref{sec:supp_eval}.
  \item \emph{4DGT Recon} To check a gain is not MoVieS-specific, we also report it under 4DGT~\cite{xu20254dgt}, a reconstructor with disjoint architecture, weights, and data.
  \item \emph{LLM judge} (Motion\,$\uparrow$, Consist.\ win\%\,$\uparrow$): a reward-blind Gemini-3.5-Flash judge, order-debiased, scoring object-motion preservation and a head-to-head consistency verdict that requires kept motion , so freezing cannot win solely. Judge details like prompts and debias mechanism can be found in Appendix~\ref{sec:supp_judge}.
  \item \emph{VideoReward}~\cite{liu2025improvingvideogenerationhuman} (VQ / MQ / TA / Overall\,$\uparrow$): paired win-rate vs.\ the distilled base under a learned video-quality model.
\end{itemize}

\subsection{Main results}
\label{sec:exp:main}

\begin{table*}[!t]
\small
\centering
\caption{\textbf{Main results on the 500 object-prominent testing set}. Metrics as defined in Sec.~\ref{sec:exp:setup}: MoVieS 4D reconstruction score; the 4DGT reconstruction~\cite{xu20254dgt} score (a reconstructor-independent cross-check, Appendix~\ref{sec:supp_eval}); the order-debiased LLM judge evaluating \emph{Motion preservation} and \emph{Consistency}); and VideoReward~\cite{liu2025improvingvideogenerationhuman} paired win\% vs.\ base.}
\label{tab:compare_short}
\setlength{\tabcolsep}{3pt}
\renewcommand{\arraystretch}{1.07}
\resizebox{\textwidth}{!}{%
\begin{tabular}{l|ccc|ccc|cc|cccc}
\toprule
\multirow{2}{*}{\textbf{Method}}
 & \multicolumn{3}{c|}{\textbf{MoVieS Recon.}}
 & \multicolumn{3}{c|}{\textbf{4DGT Recon.}}
 & \multicolumn{2}{c|}{\textbf{LLM judge}}
 & \multicolumn{4}{c}{\textbf{VideoReward win\% }} \\
\cmidrule(lr){2-4}\cmidrule(lr){5-7}\cmidrule(lr){8-9}\cmidrule(lr){10-13}
 & \textbf{PSNR}$\uparrow$ & \textbf{SSIM}$\uparrow$ & \textbf{LPIPS}$\downarrow$
 & \textbf{PSNR}$\uparrow$ & \textbf{SSIM}$\uparrow$ & \textbf{LPIPS}$\downarrow$
 & \textbf{Motion}$\uparrow$ & \textbf{Consist.}$\uparrow$
 & \textbf{VQ}$\uparrow$ & \textbf{MQ}$\uparrow$ & \textbf{TA}$\uparrow$ & \textbf{Ovr}$\uparrow$ \\
\midrule
\multicolumn{13}{c}{\textit{Reference: full-diffusion T2V baselines (non-distilled, no RL)}} \\
\midrule
Wan2.1-T2V-1.3B~\cite{wan2025wan}                                                 & 19.15 & 0.828 & 0.258 & 17.14 & 0.614 & 0.399 & ---    & ---   & ---            & ---            & ---            & --- \\
CogVideoX-1.5-5B~\cite{yang2024cogvideox}                                          & 17.02 & 0.768 & 0.279 & 15.06 & 0.518 & 0.454 & ---    & ---   & ---            & ---            & ---            & --- \\
\midrule
\multicolumn{13}{c}{\textit{Distilled AR backbone: Self-Forcing}~\cite{huang2025self} (5\,s)} \\
\midrule
Base (B0)                                                                       & 16.88 & 0.785 & 0.282 & 16.28 & 0.559 & 0.419 & ---    & ---   & ---            & ---            & ---            & --- \\
World-R1~\cite{wang2026world}                                                     & 18.52 & 0.865 & 0.197 & 16.56 & 0.566 & \textbf{0.396} & 0.832  & 75.9  & \textbf{66.0}  & 50.6           & 54.0           & 61.8 \\
VideoGPA~\cite{du2026videogpa}                                                    & 17.75 & 0.868 & 0.202 & 16.00 & 0.529 & 0.401 & 0.737  & 60.9  & 51.8           & 39.5           & \textbf{61.9}  & 48.0 \\
\rowcolor{ourscolor} $+$ \textbf{Stream4D (Ours)}                              & \textbf{20.34} & \textbf{0.874} & \textbf{0.195} & \textbf{17.24} & \textbf{0.650} & 0.417 & \textbf{0.833} & \textbf{82.2} & 62.4  & \textbf{62.8}  & 56.2           & \textbf{66.2} \\
\midrule
\multicolumn{13}{c}{\textit{Distilled AR backbone: Causal-Forcing}~\cite{zhu2026causal} (5\,s)} \\
\midrule
Base (B0)                                                                       & 15.44 & 0.767 & 0.279 & 15.03 & 0.515 & 0.450 & ---    & ---   & ---            & ---            & ---            & --- \\
World-R1~\cite{wang2026world}                                                     & 19.18 & 0.884 & 0.167 & 16.13 & 0.565 & 0.424 & 0.676  & 69.1  & 76.8           & 66.8           & 63.0           & 74.8 \\
VideoGPA~\cite{du2026videogpa}                                                    & 18.04 & 0.878 & 0.195 & 14.86 & 0.503 & 0.495 & 0.706  & 60.4  & 72.0           & 58.3           & \textbf{64.9}  & 66.3 \\
\rowcolor{ourscolor} $+$ \textbf{Stream4D (Ours)}                              & \textbf{20.97} & \textbf{0.893} & \textbf{0.150} & \textbf{17.21} & \textbf{0.643} & \textbf{0.421} & \textbf{0.765}  & \textbf{73.9} & \textbf{77.8}  & \textbf{73.2}  & 59.0           & \textbf{76.0} \\
\midrule
\multicolumn{13}{c}{\textit{Distilled AR backbone: LongLive}~\cite{yang2025longlive} (10.3\,s)} \\
\midrule
Base (B0)                                                                       & 17.44 & 0.844 & 0.231 & 15.45 & 0.521 & 0.428 & ---    & ---   & ---            & ---            & ---            & --- \\
World-R1~\cite{wang2026world}                                                     & 22.64 & \textbf{0.941} & \textbf{0.135} & 17.55 & 0.641 & 0.420 & 0.498   & 54.0  & 81.4           & 67.6           & 66.8           & 78.2 \\
VideoGPA~\cite{du2026videogpa}                                                    & 20.50 & 0.933 & 0.147 & 17.19 & 0.608 & \textbf{0.374} & 0.306   & 32.9  & 59.3           & 47.6           & 63.5           & 57.9 \\
\rowcolor{ourscolor} $+$ \textbf{Stream4D (Ours)}                              & \textbf{24.20} & 0.905 & 0.146 & \textbf{20.03} & \textbf{0.663} & 0.392 & \textbf{0.706}  & \textbf{74.2} & \textbf{81.6}  & \textbf{79.0}  & \textbf{74.0}  & \textbf{84.4} \\
\bottomrule
\end{tabular}
}
\vspace{-2mm}
\end{table*}

Table~\ref{tab:compare_short} reports Stream4D at ckpt-150 on the 500 motion-prominent testing set against the distilled base, World-R1~\cite{wang2026world} and VideoGPA~\cite{du2026videogpa} on each of the three bacbones.

\noindent\textbf{4D-consistency gains transfer across all three backbones.}
Stream4D lifts 4D-PSNR by $+3.46$\,dB on Self-Forcing , $+5.53$\,dB on Causal-Forcing, and $+6.76$\,dB on LongLive , with SSIM and LPIPS improving on all three. Under 4DGT~\cite{xu20254dgt}, a reconstructor with disjoint architecture, weights, and training data from MoVieS, Stream4D again posts the best PSNR and SSIM in every backbone block, leading World-R1 by $+0.7$ / $+1.1$ / $+2.5$\,dB. This indicates the 4D-reconstruction gain is not specific to MoVieS' model; it does not by itself certify metric-accurate geometry, a caveat we make precise in Appendix~\ref{sec:supp_eval}. Furthermore, the vision-LLM judge gives Stream4D the best consistency score on every backbone, beating base on $82.2\,\%$ / $73.9\,\%$ / $74.2\,\%$ of prompts against World-R1's $75.9\,\%$ / $69.1\,\%$ / $54.0\,\%$, confirming the improved consistency.

\noindent\textbf{Object motion is preserved; 3D method collapses it.} The vision-LLM judge scores Stream4D's motion preservation at $0.83$ / $0.77$ / $0.71$ on SF / CF / LL highest across the three backnones compared to the baselines. One can see qualitative filmstrips for all three backbones in Appendix Fig.~\ref{fig:qualitative}, showing the failure mode directly: the static-reward baselines lock the subject in place while Stream4D keeps it moving.

\noindent\textbf{Stream4D beats World-R1 on VideoReward-Overall on every backbone.}   
Overall win-rates against base are $66.2\,\%$ / $76.0\,\%$ / $84.4\,\%$, versus World-R1's $61.8\,\%$ / $74.8\,\%$ / $78.2\,\%$. The margin is largest on the motion-quality head ($+12.2$ / $+6.4$ / $+11.4$\,pp): on motion-prominent prompts, a reward that keeps motion alive wins preference over one that freezes it. 

\noindent\textbf{Human evaluation.}
We conduct a blinded human study on the 10.3s LongLive backbone, where accumulated drift and the static-collapse shortcut are most severe. We sample $50$ high-motion prompts and form the three pairings (Stream4D vs.\ base, World-R1, VideoGPA), giving $150$ two-alternative forced-choice comparisons, each judged once and distributed across $5$ raters. Each trial asks which clip shows \emph{more natural motion} and which is \emph{better consistency} jointly weighing motion amount, motion quality, and object consistency. As shown in Table~\ref{tab:human_study}, humans prefer Stream4D over both geometry-reward baselines by a wide margin---$76\,\%$ overall vs.\ World-R1 and $80\,\%$ vs.\ VideoGPA. Full protocol and screening are in Appendix~\ref{sec:supp_human}.

\begin{table}[t]
\small
\centering
\caption{\textbf{Human study on LongLive.} We focus the human evaluation on the long-horizon LongLive backbone, where accumulated drift and the static-collapse shortcut matter most. For each pair raters pick which video shows \emph{more natural motion} and which is \emph{better overall} (jointly weighing motion amount, motion quality, and object consistency); win\% is Stream4D's rate (ties count $\tfrac12$, ${>}50\%$ prefers ours). For reference we list the Gemini judge's \emph{Motion} and \emph{Consistency} win\% on the same $50$ prompts: the judge agrees with humans on every ranking (Stream4D preferred over both reward baselines, and below $50\%$ on raw motion vs.\ base), corroborating that the automatic scores are not an artifact, though the judge is somewhat more generous to ours in absolute terms.}
\label{tab:human_study}
\setlength{\tabcolsep}{6pt}
\renewcommand{\arraystretch}{1.1}
\begin{tabular}{l|cc|cc}
\toprule
\multirow{2}{*}{\textbf{Stream4D vs.}} & \multicolumn{2}{c|}{\textbf{Human win\%}$\uparrow$} & \multicolumn{2}{c}{\textbf{Gemini judge win\%}$\uparrow$} \\
\cmidrule(lr){2-3}\cmidrule(lr){4-5}
 & \textbf{Motion} & \textbf{Overall} & \textbf{Motion} & \textbf{Consist.} \\
\midrule
Base                            & $43$ & $60$ & $63$ & $70$ \\
World-R1~\cite{wang2026world}   & $72$ & $76$ & $93$ & $68$ \\
VideoGPA~\cite{du2026videogpa} & $87$ & $80$ & $95$ & $82$ \\
\bottomrule
\end{tabular}
\vspace{-1mm}
\\[2pt]
\end{table}

\subsection{Reward ablation study}
\label{sec:exp:ablation}

\begin{table*}[!t]
\small
\centering
\caption{\textbf{Reward-axis ablation:} dropping one of the three reward axes, on all three backbones. Each block lists the distilled base, the deployed Stream4D recipe, then the three single-axis drops. Metric conventions match Table~\ref{tab:compare_short}.}
\label{tab:ablation_axes}
\setlength{\tabcolsep}{3.5pt}
\renewcommand{\arraystretch}{1.05}
\resizebox{\textwidth}{!}{%
\begin{tabular}{l|ccc|ccc|cc|cccc}
\toprule
\multirow{2}{*}{\textbf{Reward configuration}}
 & \multicolumn{3}{c|}{\textbf{MoVieS Recon.}}
 & \multicolumn{3}{c|}{\textbf{4DGT Recon.}}
 & \multicolumn{2}{c|}{\textbf{LLM judge}}
 & \multicolumn{4}{c}{\textbf{VideoReward win\%}} \\
\cmidrule(lr){2-4}\cmidrule(lr){5-7}\cmidrule(lr){8-9}\cmidrule(lr){10-13}
 & \textbf{PSNR}$\uparrow$ & \textbf{SSIM}$\uparrow$ & \textbf{LPIPS}$\downarrow$
 & \textbf{PSNR}$\uparrow$ & \textbf{SSIM}$\uparrow$ & \textbf{LPIPS}$\downarrow$
 & \textbf{Motion}$\uparrow$ & \textbf{Consist.}$\uparrow$
 & \textbf{VQ}$\uparrow$ & \textbf{MQ}$\uparrow$ & \textbf{TA}$\uparrow$ & \textbf{Ovr}$\uparrow$ \\
\midrule
\multicolumn{13}{c}{\textit{Backbone: Self-Forcing (5\,s)}} \\
\midrule
Base (no RL)                                     & 16.88 & 0.785 & 0.282 & 16.28 & 0.559 & 0.419 & ---   & ---   & ---   & ---   & ---   & --- \\
\rowcolor{ourscolor} \textbf{Stream4D (Ours)} & 20.34 & 0.874 & 0.195 & 17.24 & 0.650 & 0.417 & 0.833 & 82.2  & 62.4  & 62.8  & 56.2  & 66.2 \\
\quad$-$ perceptual anchor (a)                   & 22.39 & 0.905 & 0.163 & 18.53 & 0.723 & 0.371 & 0.830 & 81.7  & 66.4  & 64.8  & 52.8  & 66.6 \\
\quad$-$ motion term (b)                         & 24.37 & 0.952 & 0.096 & 19.36 & 0.738 & 0.328 & 0.341 & 31.1  & 63.0  & 56.8  & 44.0  & 59.2 \\
\quad$-$ reconstruction (c)                      & 13.98 & 0.692 & 0.342 & 14.17 & 0.466 & 0.476 & 0.983 & 24.2  & 31.6  & 26.4  & 34.6  & 28.2 \\
\midrule
\multicolumn{13}{c}{\textit{Backbone: Causal-Forcing (5\,s)}} \\
\midrule
Base (no RL)                                     & 15.44 & 0.767 & 0.279 & 15.03 & 0.515 & 0.450 & ---   & ---   & ---   & ---   & ---   & --- \\
\rowcolor{ourscolor} \textbf{Stream4D (Ours)} & 20.97 & 0.893 & 0.150 & 17.21 & 0.643 & 0.421 & 0.765 & 73.9  & 77.8  & 73.2  & 59.0  & 76.0 \\
\quad$-$ perceptual anchor (a)                   & 23.93 & 0.901 & 0.106 & 18.48 & 0.792 & 0.341 & 0.591 & 47.3  & 67.0  & 57.8  & 48.8  & 64.4 \\
\quad$-$ motion term (b)$^{\dagger}$             & 24.19 & 0.862 & 0.094 & 18.89 & 0.776 & 0.329 & 0.453 & 22.4  & 71.6  & 56.2  & 51.0  & 63.4 \\
\quad$-$ reconstruction (c)                      & 14.88 & 0.706 & 0.297 & 14.41 & 0.414 & 0.495 & 0.984 & 45.1  & 45.2  & 50.4  & 50.0  & 49.2 \\
\midrule
\multicolumn{13}{c}{\textit{Backbone: LongLive (10.3\,s)}} \\
\midrule
Base (no RL)                                     & 17.44 & 0.844 & 0.231 & 15.45 & 0.521 & 0.428 & ---   & ---   & ---   & ---   & ---   & --- \\
\rowcolor{ourscolor} \textbf{Stream4D (Ours)} & 24.20 & 0.905 & 0.146 & 20.03 & 0.663 & 0.392 & 0.706 & 74.2  & 81.6  & 79.0  & 74.0  & 84.4 \\
\quad$-$ perceptual anchor (a)                   & 20.57 & 0.936 & 0.143 & 16.84 & 0.639 & 0.379 & 0.745 & 74.2  & 75.0  & 60.0  & 57.0  & 69.4 \\
\quad$-$ motion term (b)                         & 22.84 & 0.931 & 0.133 & 17.69 & 0.611 & 0.438 & 0.416 & 44.6  & 78.4  & 66.6  & 65.4  & 74.4 \\
\quad$-$ reconstruction (c)                      & 18.29 & 0.844 & 0.217 & 16.13 & 0.520 & 0.427 & 0.956 & 80.2  & 54.2  & 66.8  & 59.8  & 63.4 \\
\bottomrule
\end{tabular}
}
\vspace{-2mm}
\end{table*}

We study the reward in two parts: dropping each of the three axes in Table~\ref{tab:ablation_axes}, and varying the shape of the motion term in Table~\ref{tab:ablation_motion}.

\noindent\textbf{Each axis is necessary on at least one backbone; none can be dropped safely on all three.}
In Table~\ref{tab:ablation_axes}, dropping the \emph{motion term} (b) collapses motion on every backbone and loses the vlm consistency score to base on all three despite near-best PSNR. Dropping \emph{reconstruction} (c) can achieve the most motion, but coherence collapses. Dropping the \emph{perceptual anchor} makes CF lose to base on vlm consistency judge and on LL it costs $3.6$\,dB of 4D-PSNR and $15$\,pp of VideoReward-Overall.

\begin{wrapfigure}{r}{0.50\textwidth}
\vspace{-5mm}
\centering
\includegraphics[width=0.44\textwidth]{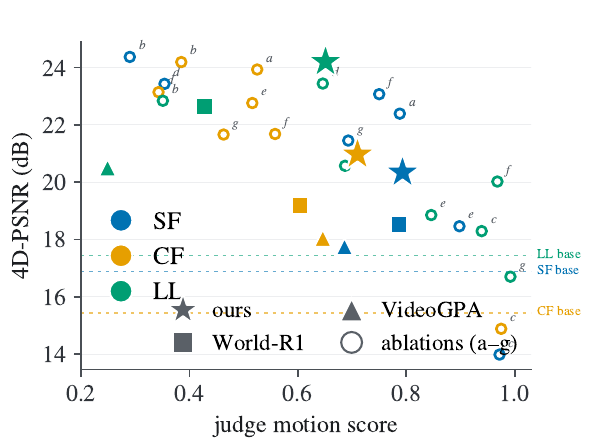}
\caption{\textbf{
Reconstruction-motion trade-off}. Deployed recipes (stars) sit on the upper-right frontier.}
\label{fig:analysis}
\vspace{-4mm}
\end{wrapfigure}

\begin{table*}[!t]
\small
\centering
\caption{\textbf{Motion-term formulation ablation:} varying the shape of the motion term $R_{\text{mot}}$ while keeping the reconstruction and perceptual axes fixed, on all three backbones.}
\label{tab:ablation_motion}
\setlength{\tabcolsep}{3.5pt}
\renewcommand{\arraystretch}{1.05}
\resizebox{\textwidth}{!}{%
\begin{tabular}{l|ccc|ccc|cc|cccc}
\toprule
\multirow{2}{*}{\textbf{Motion-term $R_{\text{mot}}$}}
 & \multicolumn{3}{c|}{\textbf{MoVieS Recon.}}
 & \multicolumn{3}{c|}{\textbf{4DGT Recon.}}
 & \multicolumn{2}{c|}{\textbf{LLM judge}}
 & \multicolumn{4}{c}{\textbf{VideoReward win\%}} \\
\cmidrule(lr){2-4}\cmidrule(lr){5-7}\cmidrule(lr){8-9}\cmidrule(lr){10-13}
 & \textbf{PSNR}$\uparrow$ & \textbf{SSIM}$\uparrow$ & \textbf{LPIPS}$\downarrow$
 & \textbf{PSNR}$\uparrow$ & \textbf{SSIM}$\uparrow$ & \textbf{LPIPS}$\downarrow$
 & \textbf{Motion}$\uparrow$ & \textbf{Consist.}$\uparrow$
 & \textbf{VQ}$\uparrow$ & \textbf{MQ}$\uparrow$ & \textbf{TA}$\uparrow$ & \textbf{Ovr}$\uparrow$ \\
\midrule
\multicolumn{13}{c}{\textit{Backbone: Self-Forcing (5\,s)}} \\
\midrule
Base (no RL)                                        & 16.88 & 0.785 & 0.282 & 16.28 & 0.559 & 0.419 & ---   & ---   & ---   & ---   & ---   & --- \\
\rowcolor{ourscolor} \textbf{$g(m)\!\cdot\!\mathrm{smooth}\!\cdot\!\mathrm{rigid}$ } & 20.34 & 0.874 & 0.195 & 17.24 & 0.650 & 0.417 & 0.833 & 82.2  & 62.4  & 62.8  & 56.2  & 66.2 \\
\quad$\mathrm{smooth}\!\cdot\!\mathrm{rigid}$,             & 23.43 & 0.947 & 0.119 & 18.02 & 0.671 & 0.405 & 0.406 & 36.4  & 78.4  & 60.4  & 53.8  & 74.6 \\
\quad $m\!\cdot\!\mathrm{smooth}\!\cdot\!\mathrm{rigid}$ (e) & 18.46 & 0.833 & 0.234 & 16.85 & 0.632 & 0.392 & 0.923 & 71.2  & 46.4  & 47.4  & 41.8  & 46.0 \\
\quad $g(m)$ only                    & 23.07 & 0.907 & 0.155 & 18.99 & 0.745 & 0.356 & 0.795 & 75.8  & 64.4  & 65.0  & 52.2  & 64.0 \\
\quad $g(m)\!+\!\mathrm{smooth}\!+\!\mathrm{rigid}$                             & 21.45 & 0.907 & 0.165 & 17.37 & 0.647 & 0.410 & 0.750 & 75.9  & 65.8  & 57.0  & 56.0  & 65.6 \\
\midrule
\multicolumn{13}{c}{\textit{Backbone: Causal-Forcing (5\,s)}} \\
\midrule
Base (no RL)                                        & 15.44 & 0.767 & 0.279 & 15.03 & 0.515 & 0.450 & ---   & ---   & ---   & ---   & ---   & --- \\
\rowcolor{ourscolor} \textbf{$g(m)\!\cdot\!\mathrm{smooth}\!\cdot\!\mathrm{rigid}$ } & 20.97 & 0.893 & 0.150 & 17.21 & 0.643 & 0.421 & 0.765 & 73.9  & 77.8  & 73.2  & 59.0  & 76.0 \\
\quad$\mathrm{smooth}\!\cdot\!\mathrm{rigid}$,             & 23.14 & 0.922 & 0.116 & 18.37 & 0.741 & 0.338 & 0.396 & 31.2  & 75.0  & 65.8  & 51.0  & 71.0 \\
\quad $m\!\cdot\!\mathrm{smooth}\!\cdot\!\mathrm{rigid}$ (e) & 22.76 & 0.918 & 0.120 & 17.74 & 0.728 & 0.400 & 0.578 & 61.8  & 69.6  & 68.6  & 49.6  & 68.8 \\
\quad $g(m)$ only                    & 21.68 & 0.898 & 0.130 & 17.85 & 0.710 & 0.355 & 0.619 & 58.4  & 72.2  & 64.0  & 50.4  & 68.2 \\
\quad $g(m)\!+\!\mathrm{smooth}\!+\!\mathrm{rigid}$                             & 21.66 & 0.886 & 0.131 & 18.16 & 0.685 & 0.343 & 0.525 & 44.4  & 74.2  & 61.8  & 51.8  & 68.0 \\
\midrule
\multicolumn{13}{c}{\textit{Backbone: LongLive (10.3\,s)}} \\
\midrule
Base (no RL)                                        & 17.44 & 0.844 & 0.231 & 15.45 & 0.521 & 0.428 & ---   & ---   & ---   & ---   & ---   & --- \\
\rowcolor{ourscolor} \textbf{$g(m)\!\cdot\!\mathrm{smooth}\!\cdot\!\mathrm{rigid}$} & 24.20 & 0.905 & 0.146 & 20.03 & 0.663 & 0.392 & 0.706 & 74.2  & 81.6  & 79.0  & 74.0  & 84.4 \\
\quad$\mathrm{smooth}\!\cdot\!\mathrm{rigid}$,             & 23.44 & 0.924 & 0.142 & 18.58 & 0.652 & 0.417 & 0.699 & 74.1  & 74.6  & 72.2  & 72.6  & 79.2 \\
\quad  $m\!\cdot\!\mathrm{smooth}\!\cdot\!\mathrm{rigid}$ (e) & 18.85 & 0.889 & 0.193 & 15.91 & 0.578 & 0.459 & 0.884 & 58.6  & 58.0  & 56.8  & 60.4  & 60.6 \\
\quad $g(m)$ only                    & 20.02 & 0.847 & 0.232 & 18.92 & 0.766 & 0.287 & 0.978 & 64.0  & 37.4  & 64.0  & 50.6  & 50.2 \\
\quad $g(m)\!+\!\mathrm{smooth}\!+\!\mathrm{rigid}$                             & 16.70 & 0.763 & 0.327 & 16.40 & 0.700 & 0.346 & 0.995 & 30.6  & 30.6  & 48.2  & 41.4  & 38.0 \\
\bottomrule
\end{tabular}
}
\vspace{-2mm}
\end{table*}

\noindent\textbf{The motion term needs its peaked, multiplicative shape.}
In Table~\ref{tab:ablation_motion}, removing the gate (d) freezes SF and CF as smooth and rigid are trivially maximized by a near-static rollout. A linear term (e) over-drives motion on SF/LL at a reconstruction and preference cost. The gate alone (f) can preserve the motion and improve the reconstruction, but it will cost the VideoReward 2.2 7.8, and 34.2 pp and is less preferred by the vlm consistency judge, indicating the low quality motion appears without these two terms. The additive split (g) is benign on SF but a net loss on CF and catastrophic on LL. 

\noindent\textbf{Trade-off between motion and reconstruction.} Figure~\ref{fig:analysis} plots every trained variant in the reconstruction-motion plane. We can observe two failure modes: high-recon/low-motion frozen runs (upper left, e.g.\ rows (b),(d)) and high-motion/low-recon incoherent runs (lower right, e.g.\ row (c)). Note that the deployed recipes (stars) sit on the upper-right frontier.

\noindent\textbf{Sensitivity to the gate target $m_{\text{nat}}$.}
The one constant the recipe calibrates from data is the gate target $m_{\text{nat}}\!=\!0.020$. Table~\ref{tab:mnat_sensitivity} ablates it at 0.010 and 0.030 on Self-Forcing and shows the gate behaves as designed: motion rises monotonically from $0.52 \to 0.83 \to 0.92$ and trades off MoVieS-PSNR from $21.10 \to 20.34 \to 18.87$. Setting it as 0.020 just hits a balance between motion and consistency, gaining the best LLM judge Consist score and Video reward results.
\begin{table*}[!t]
\small
\centering
\caption{\textbf{Sensitivity to the gate target $m_{\text{nat}}$} on Self-Forcing. The $0.020$ row is the deployed recipe, landing on the knee of the motion-fidelity trade-off, taking the best Consistency and VideoReward-Overall.}
\label{tab:mnat_sensitivity}
\setlength{\tabcolsep}{3pt}
\renewcommand{\arraystretch}{1.07}
\resizebox{\textwidth}{!}{%
\begin{tabular}{l|ccc|ccc|cc|cccc}
\toprule
\multirow{2}{*}{\textbf{$m_{\text{nat}}$}}
 & \multicolumn{3}{c|}{\textbf{MoVieS Recon.}}
 & \multicolumn{3}{c|}{\textbf{4DGT Recon.}}
 & \multicolumn{2}{c|}{\textbf{LLM judge}}
 & \multicolumn{4}{c}{\textbf{VideoReward win\% vs base}} \\
\cmidrule(lr){2-4}\cmidrule(lr){5-7}\cmidrule(lr){8-9}\cmidrule(lr){10-13}
 & \textbf{PSNR}$\uparrow$ & \textbf{SSIM}$\uparrow$ & \textbf{LPIPS}$\downarrow$
 & \textbf{PSNR}$\uparrow$ & \textbf{SSIM}$\uparrow$ & \textbf{LPIPS}$\downarrow$
 & \textbf{Motion}$\uparrow$ & \textbf{Consist.}$\uparrow$
 & \textbf{VQ}$\uparrow$ & \textbf{MQ}$\uparrow$ & \textbf{TA}$\uparrow$ & \textbf{Ovr}$\uparrow$ \\
\midrule
Self-Forcing base                                & 16.88 & 0.785 & 0.282 & 16.28 & 0.559 & 0.419 & ---   & ---   & ---   & ---   & ---   & --- \\
\midrule
$0.010$ ($0.5\times$)                             & \textbf{21.10} & \textbf{0.924} & \textbf{0.135} & \textbf{17.86} & \textbf{0.672} & \textbf{0.345} & 0.522 & 44.7  & \textbf{62.8} & 59.2  & \textbf{65.6} & 63.8 \\
\rowcolor{ourscolor} $\mathbf{0.020}$ \textbf{(deployed)} & 20.34 & 0.874 & 0.195 & 17.24 & 0.650 & 0.417 & 0.833 & \textbf{82.2} & 62.4  & \textbf{62.8} & 56.2  & \textbf{66.2} \\
$0.030$ ($1.5\times$)                             & 18.87 & 0.836 & 0.228 & 17.50 & 0.645 & 0.374 & \textbf{0.916} & 74.4  & 54.0  & 51.0  & 45.6  & 53.2 \\
\bottomrule
\end{tabular}
}
\vspace{-2mm}
\end{table*}

\section{Conclusion}
\label{sec:conclusion}

We presented Stream4D, a reinforcement-learning recipe that moves geometric reward design for streaming AR video models from 3D- to 4D-consistency. A rigid 3D reconstruction cannot model a dynamic scene, so it penalizes motion and lets the policy earn reward by freezing the scene; a feed-forward 4D reconstruction does not penalize motion, and paired with a peaked motion gate and a light perceptual anchor it instead rewards coherent \emph{moving} content. The single recipe transfers across three distilled backbones, improving 4D reconstruction, motion preservation, and preference win-rate over both the distilled base and a faithful World-R1 reproduction, with a reconstructor-independent cross-check and a reward-blind judge corroborating each gain. A natural next step is a streaming 4D reconstructor to match LongLive's native horizon, and grading rollouts against explicit action or camera inputs for embodied control.

\bibliographystyle{plainnat}
\bibliography{main}

@String(CVPR  = {IEEE Conf. Comput. Vis. Pattern Recog.})

@String(ECCV  = {Eur. Conf. Comput. Vis.})

@String(NeurIPS = {Adv. Neural Inform. Process. Syst.})

@String(ICML  = {Int. Conf. Mach. Learn.})

@String(ICLR  = {Int. Conf. Learn. Represent.})

@String(CVPR  = {CVPR})

@String(ECCV  = {ECCV})

@String(NeurIPS = {NeurIPS})

@String(ICML  = {ICML})

@String(ICLR  = {ICLR})

@inproceedings{yin2025slow,
  title={From slow bidirectional to fast autoregressive video diffusion models},
  author={Yin, Tianwei and Zhang, Qiang and Zhang, Richard and Freeman, William T and Durand, Fredo and Shechtman, Eli and Huang, Xun},
  booktitle={Proceedings of the Computer Vision and Pattern Recognition Conference},
  pages={22963--22974},
  year={2025}
}

@article{huang2025self,
  title={Self Forcing: Bridging the Train-Test Gap in Autoregressive Video Diffusion},
  author={Huang, Xun and Li, Zhengqi and He, Guande and Zhou, Mingyuan and Shechtman, Eli},
  journal={arXiv preprint arXiv:2506.08009},
  year={2025}
}

@inproceedings{yin2024one,
  title={One-step diffusion with distribution matching distillation},
  author={Yin, Tianwei and Gharbi, Micha{\"e}l and Zhang, Richard and Shechtman, Eli and Durand, Fredo and Freeman, William T and Park, Taesung},
  booktitle={Proceedings of the IEEE/CVF conference on computer vision and pattern recognition},
  pages={6613--6623},
  year={2024}
}

@article{wan2025wan,
  title={Wan: Open and advanced large-scale video generative models},
  author={Wan, Team and Wang, Ang and Ai, Baole and Wen, Bin and Mao, Chaojie and Xie, Chen-Wei and Chen, Di and Yu, Feiwu and Zhao, Haiming and Yang, Jianxiao and others},
  journal={arXiv preprint arXiv:2503.20314},
  year={2025}
}

@article{chen2024diffusion,
  title={Diffusion forcing: Next-token prediction meets full-sequence diffusion},
  author={Chen, Boyuan and Mart{\'\i} Mons{\'o}, Diego and Du, Yilun and Simchowitz, Max and Tedrake, Russ and Sitzmann, Vincent},
  journal={Advances in Neural Information Processing Systems},
  volume={37},
  pages={24081--24125},
  year={2024}
}

@article{wang2024vidprom,
  title={Vidprom: A million-scale real prompt-gallery dataset for text-to-video diffusion models},
  author={Wang, Wenhao and Yang, Yi},
  journal={Advances in Neural Information Processing Systems},
  volume={37},
  pages={65618--65642},
  year={2024}
}

@article{kong2024hunyuanvideo,
  title={Hunyuanvideo: A systematic framework for large video generative models},
  author={Kong, Weijie and Tian, Qi and Zhang, Zijian and Min, Rox and Dai, Zuozhuo and Zhou, Jin and Xiong, Jiangfeng and Li, Xin and Wu, Bo and Zhang, Jianwei and others},
  journal={arXiv preprint arXiv:2412.03603},
  year={2024}
}

@article{bao2024vidu,
  title={Vidu: a highly consistent, dynamic and skilled text-to-video generator with diffusion models},
  author={Bao, Fan and Xiang, Chendong and Yue, Gang and He, Guande and Zhu, Hongzhou and Zheng, Kaiwen and Zhao, Min and Liu, Shilong and Wang, Yaole and Zhu, Jun},
  journal={arXiv preprint arXiv:2405.04233},
  year={2024}
}

@article{yang2024cogvideox,
  title={Cogvideox: Text-to-video diffusion models with an expert transformer},
  author={Yang, Zhuoyi and Teng, Jiayan and Zheng, Wendi and Ding, Ming and Huang, Shiyu and Xu, Jiazheng and Yang, Yuanming and Hong, Wenyi and Zhang, Xiaohan and Feng, Guanyu and others},
  journal={arXiv preprint arXiv:2408.06072},
  year={2024}
}

@article{yang2025longlive,
  title={Longlive: Real-time interactive long video generation},
  author={Yang, Shuai and Huang, Wei and Chu, Ruihang and Xiao, Yicheng and Zhao, Yuyang and Wang, Xianbang and Li, Muyang and Xie, Enze and Chen, Yingcong and Lu, Yao and others},
  journal={arXiv preprint arXiv:2509.22622},
  year={2025}
}

@article{cui2025self,
  title={Self-forcing++: Towards minute-scale high-quality video generation},
  author={Cui, Justin and Wu, Jie and Li, Ming and Yang, Tao and Li, Xiaojie and Wang, Rui and Bai, Andrew and Ban, Yuanhao and Hsieh, Cho-Jui},
  journal={arXiv preprint arXiv:2510.02283},
  year={2025}
}

@article{sun2025worldplay,
  title={WorldPlay: Towards Long-Term Geometric Consistency for Real-Time Interactive World Modeling},
  author={Sun, Wenqiang and Zhang, Haiyu and Wang, Haoyuan and Wu, Junta and Wang, Zehan and Wang, Zhenwei and Wang, Yunhong and Zhang, Jun and Wang, Tengfei and Guo, Chunchao},
  journal={arXiv preprint arXiv:2512.14614},
  year={2025}
}

@article{liu2025flowgrpo,
  title={Flow-grpo: Training flow matching models via online rl},
  author={Liu, Jie and Liu, Gongye and Liang, Jiajun and Li, Yangguang and Liu, Jiaheng and Wang, Xintao and Wan, Pengfei and Zhang, Di and Ouyang, Wanli},
  journal={arXiv preprint arXiv:2505.05470},
  year={2025}
}

@article{zheng2025diffusionnft,
  title={Diffusionnft: Online diffusion reinforcement with forward process},
  author={Zheng, Kaiwen and Chen, Huayu and Ye, Haotian and Wang, Haoxiang and Zhang, Qinsheng and Jiang, Kai and Su, Hang and Ermon, Stefano and Zhu, Jun and Liu, Ming-Yu},
  journal={arXiv preprint arXiv:2509.16117},
  year={2025}
}

@article{hu2024acdit,
  title={Acdit: Interpolating autoregressive conditional modeling and diffusion transformer},
  author={Hu, Jinyi and Hu, Shengding and Song, Yuxuan and Huang, Yufei and Wang, Mingxuan and Zhou, Hao and Liu, Zhiyuan and Ma, Wei-Ying and Sun, Maosong},
  journal={arXiv preprint arXiv:2412.07720},
  year={2024}
}

@article{gao2024ca2,
  title={Ca2-vdm: Efficient autoregressive video diffusion model with causal generation and cache sharing},
  author={Gao, Kaifeng and Shi, Jiaxin and Zhang, Hanwang and Wang, Chunping and Xiao, Jun and Chen, Long},
  journal={arXiv preprint arXiv:2411.16375},
  year={2024}
}

@article{xue2025dancegrpo,
  title={Dancegrpo: Unleashing grpo on visual generation},
  author={Xue, Zeyue and Wu, Jie and Gao, Yu and Kong, Fangyuan and Zhu, Lingting and Chen, Mengzhao and Liu, Zhiheng and Liu, Wei and Guo, Qiushan and Huang, Weilin and others},
  journal={arXiv preprint arXiv:2505.07818},
  year={2025}
}

@inproceedings{zhu2026causal,
  title={Causal Forcing: Autoregressive Diffusion Distillation Done Right for High-Quality Real-Time Interactive Video Generation},
  author={Zhu, Hongzhou and Zhao, Min and He, Guande and Su, Hang and Li, Chongxuan and Zhu, Jun},
  booktitle={International Conference on Machine Learning (ICML)},
  note={arXiv:2602.02214},
  year={2026}
}

@article{wu2023hpsv2,
  title={Human Preference Score v2: A Solid Benchmark for Evaluating Human Preferences of Text-to-Image Synthesis},
  author={Wu, Xiaoshi and Hao, Yiming and Sun, Keqiang and Chen, Yixiong and Zhu, Feng and Zhao, Rui and Li, Hongsheng},
  journal={arXiv preprint arXiv:2306.09341},
  year={2023}
}

@inproceedings{yesiltepe2025infinityrope,
  title={Infinity-rope: Action-controllable infinite video generation emerges from autoregressive self-rollout},
  author={Yesiltepe, Hidir and Meral, Tuna Han Salih and Akan, Adil Kaan and Oktay, Kaan and Yanardag, Pinar},
  booktitle={CVPR},
  note={arXiv:2511.20649},
  year={2026}
}

@article{he2025gardo,
  title={GARDO: Reinforcing Diffusion Models without Reward Hacking},
  author={He, Haoran and Ye, Yuxiao and Liu, Jie and Liang, Jiajun and Wang, Zhiyong and Yuan, Ziyang and Wang, Xintao and Mao, Hangyu and Wan, Pengfei and Pan, Ling},
  journal={arXiv preprint arXiv:2512.24138},
  year={2025}
}

@article{wang2026worldcompass,
  title={WorldCompass: Reinforcement Learning for Long-Horizon World Models},
  author={Wang, Zehan and Wang, Tengfei and Zhang, Haiyu and Zuo, Xuhui and Wu, Junta and Wang, Haoyuan and Sun, Wenqiang and Wang, Zhenwei and Cao, Chenjie and Zhao, Hengshuang and others},
  journal={arXiv preprint arXiv:2602.09022},
  year={2026}
}

@article{lin2025depth,
  title={Depth anything 3: Recovering the visual space from any views},
  author={Lin, Haotong and Chen, Sili and Liew, Junhao and Chen, Donny Y and Li, Zhenyu and Shi, Guang and Feng, Jiashi and Kang, Bingyi},
  journal={arXiv preprint arXiv:2511.10647},
  year={2025}
}

@article{ye2025gsplat,
  title={gsplat: An open-source library for Gaussian splatting},
  author={Ye, Vickie and Li, Ruilong and Kerr, Justin and Turkulainen, Matias and Yi, Brent and Pan, Zhuoyang and Seiskari, Otto and Ye, Jianbo and Hu, Jeffrey and Tancik, Matthew and others},
  journal={Journal of Machine Learning Research},
  volume={26},
  number={34},
  pages={1--17},
  year={2025}
}

@article{Qwen3-VL,
  title={Qwen3-VL Technical Report},
  author={Shuai Bai and Yuxuan Cai and Ruizhe Chen and Keqin Chen and Xionghui Chen and Zesen Cheng and Lianghao Deng and Wei Ding and Chang Gao and Chunjiang Ge and Wenbin Ge and Zhifang Guo and Qidong Huang and Jie Huang and Fei Huang and Binyuan Hui and Shutong Jiang and Zhaohai Li and Mingsheng Li and Mei Li and Kaixin Li and Zicheng Lin and Junyang Lin and Xuejing Liu and Jiawei Liu and Chenglong Liu and Yang Liu and Dayiheng Liu and Shixuan Liu and Dunjie Lu and Ruilin Luo and Chenxu Lv and Rui Men and Lingchen Meng and Xuancheng Ren and Xingzhang Ren and Sibo Song and Yuchong Sun and Jun Tang and Jianhong Tu and Jianqiang Wan and Peng Wang and Pengfei Wang and Qiuyue Wang and Yuxuan Wang and Tianbao Xie and Yiheng Xu and Haiyang Xu and Jin Xu and Zhibo Yang and Mingkun Yang and Jianxin Yang and An Yang and Bowen Yu and Fei Zhang and Hang Zhang and Xi Zhang and Bo Zheng and Humen Zhong and Jingren Zhou and Fan Zhou and Jing Zhou and Yuanzhi Zhu and Ke Zhu},
  journal={arXiv preprint arXiv:2511.21631},
  year={2025}
}

@inproceedings{liu2025improvingvideogenerationhuman,
  title={Improving Video Generation with Human Feedback},
  author={Liu, Jie and Liu, Gongye and Liang, Jiajun and Yuan, Ziyang and Liu, Xiaokun and Zheng, Mingwu and Wu, Xiele and Wang, Qiulin and Xia, Menghan and Wang, Xintao and Liu, Xiaohong and Yang, Fei and Wan, Pengfei and Zhang, Di and Gai, Kun and Yang, Yujiu and Ouyang, Wanli},
  booktitle={Advances in Neural Information Processing Systems},
  year={2025}
}

@article{kupyn2025epipolar,
  title={Epipolar Geometry Improves Video Generation Models},
  author={Kupyn, Orest and Manhardt, Fabian and Tombari, Federico and Rupprecht, Christian},
  journal={arXiv preprint arXiv:2510.21615},
  year={2025}
}

@article{astrolabe_2026,
  title={Astrolabe: Steering Forward-Process Reinforcement Learning for Distilled Autoregressive Video Models},
  author={Zhang, Songchun and Xue, Zeyue and Fu, Siming and Huang, Jie and Kong, Xianghao and Ma, Yue and Huang, Haoyang and Duan, Nan and Rao, Anyi},
  journal={arXiv preprint arXiv:2603.17051},
  year={2026},
  note={ECCV 2026 submission}
}

@inproceedings{wang2026world,
  title={World-R1: Reinforcing 3D Constraints for Text-to-Video Generation},
  author={Wang, Weijie and He, Xiaoxuan and Gu, Youping and Yang, Yifan and Zhang, Zeyu and He, Yefei and Ding, Yanbo and Hu, Xirui and Chen, Donny Y and He, Zhiyuan and others},
  booktitle={International Conference on Machine Learning (ICML)},
  note={arXiv:2604.24764},
  year={2026}
}

@inproceedings{du2026videogpa,
  title={VideoGPA: Distilling Geometry Priors for 3D-Consistent Video Generation},
  author={Du, Hongyang and Ye, Junjie and Cong, Xiaoyan and Li, Runhao and Ni, Jingcheng and Agarwal, Aman and Zhou, Zeqi and Li, Zekun and Balestriero, Randall and Wang, Yue},
  booktitle={International Conference on Machine Learning (ICML)},
  note={arXiv:2601.23286},
  year={2026}
}

@inproceedings{lin2025movies,
  title={Mo{V}ie{S}: Motion-Aware 4D Dynamic View Synthesis in One Second},
  author={Lin, Chenguo and Lin, Yuchen and Pan, Panwang and Yu, Yifan and Hu, Tao and Yan, Honglei and Fragkiadaki, Katerina and Mu, Yadong},
  booktitle={CVPR},
  year={2026},
  note={arXiv:2507.10065}
}

@inproceedings{streamvggt_2026,
  title={Streaming 4D Visual Geometry Transformer},
  author={Zhuo, Dong and Zheng, Wenzhao and Guo, Jiahe and Wu, Yuqi and Zhou, Jie and Lu, Jiwen},
  booktitle={ICLR},
  year={2026},
  note={arXiv:2507.11539}
}

@inproceedings{xu20254dgt,
  title={{4DGT}: Learning a {4D} {G}aussian Transformer Using Real-World Monocular Videos},
  author={Xu, Zhen and Li, Zhengqin and Dong, Zhao and Zhou, Xiaowei and Newcombe, Richard and Lv, Zhaoyang},
  booktitle={NeurIPS},
  year={2025},
  note={arXiv:2506.08015}
}

@article{liu2026gdpo,
  title={Gdpo: Group reward-decoupled normalization policy optimization for multi-reward rl optimization},
  author={Liu, Shih-Yang and Dong, Xin and Lu, Ximing and Diao, Shizhe and Belcak, Peter and Liu, Mingjie and Chen, Min-Hung and Yin, Hongxu and Wang, Yu-Chiang Frank and Cheng, Kwang-Ting and others},
  journal={arXiv preprint arXiv:2601.05242},
  year={2026}
}

@inproceedings{zhang2018unreasonable,
  title={The unreasonable effectiveness of deep features as a perceptual metric},
  author={Zhang, Richard and Isola, Phillip and Efros, Alexei A and Shechtman, Eli and Wang, Oliver},
  booktitle={2018 IEEE/CVF conference on computer vision and pattern recognition},
  pages={586--595},
  year={2018},
  organization={IEEE}
}

\appendix
\clearpage
\appendix
\setcounter{page}{1}
\setcounter{figure}{0}
\setcounter{table}{0}
\renewcommand{\thesection}{\Alph{section}}
\renewcommand{\thefigure}{S\arabic{figure}}
\renewcommand{\thetable}{S\arabic{table}}

This appendix is organized as follows: Appendix~\ref{sec:supp_reward} details the deployed reward; Appendix~\ref{sec:supp_hyper} lists training hyperparameters and compute; Appendix~\ref{sec:algorithm} gives the full training loop; Appendix~\ref{sec:supp_eval} details the evaluation protocol and metrics; Appendix~\ref{sec:supp_judge} documents the LLM judge (prompts, position debiasing, reproducibility); Appendix~\ref{sec:supp_human} details the human-study protocol;Appendix~\ref{sec:supp_rand500} reports the random-subset robustness check; Appendix~\ref{sec:supp_qual} collects qualitative comparisons; Appendix~\ref{sec:exp:limits} states limitations; and Appendix~\ref{sec:supp_discussion} discusses the scope of the reward backbone.

\section{Reward Implementation Details}
\label{sec:supp_reward}

This appendix gives the deployed-recipe details that the main paper Sec.~\ref{sec:method:reward} compresses. The reward $R(W^{(i)}) = w_{\text{recon}}\!\cdot\!\tilde z[R_{\text{recon}}] + w_{\text{mot}}\!\cdot\!\tilde z[R_{\text{mot}}] + w_{\text{hpsv2}}\!\cdot\!\tilde z[R_{\text{hpsv2}}]$ (Eq.~\ref{eq:reward}) is computed as follows; the per-backbone axis weights $(w_{\text{recon}}, w_{\text{mot}}, w_{\text{hpsv2}})$ are $(1.0,\,1.0,\,0.3)$ on Self-Forcing, $(1.0,\,0.5,\,0.6)$ on Causal-Forcing, and $(0.8,\,1.0,\,0.6)$ on LongLive, and all other hyperparameters are shared.

\noindent\textbf{4D-GS reconstruction backbone.}
StreamVGGT~\cite{streamvggt_2026} predicts per-frame camera extrinsics $\{C_t\}_{t=1}^{T}$ and intrinsics from a 26-frame linearly-subsampled version of the rollout at $294\!\times\!518$. MoVieS~\cite{lin2025movies} consumes the same subsampled frames and the predicted cameras and returns: (i)~a canonical Gaussian point cloud $G_{\text{can}}$ encoding frame-invariant geometry, (ii)~per-frame Gaussian-attribute overrides (opacity, scale, color) $\{A_t\}_{t=1}^{T}$, and (iii)~per-frame scene-flow offsets $\{\Delta_t\}_{t=1}^{T}$. We re-render each frame with \texttt{GaussianRenderer.render} using $G_{\text{can}}\!\oplus\!A_t\!\oplus\!\Delta_t$ at $C_t$, then compute frame-wise LPIPS-AlexNet against the input frame and combine as in Eq.~\ref{eq:r_recon}. A single MoVieS forward per candidate rollout is shared between $R_{\text{recon}}$ and $R_{\text{mot}}$ via a per-batch cache.

\noindent\textbf{Motion-quality conjunction.}
$R_{\text{mot}} = g(m)\cdot\mathrm{smooth}\cdot\mathrm{rigid}$ exactly as in Eqs.~\ref{eq:r_mot}--\ref{eq:rigid}: $g(m) = \exp(-(m\!-\!m_{\text{nat}})^2/(2\sigma^2))$ with $m_{\text{nat}}\!=\!0.020$, $\sigma\!=\!0.010$ (MoVieS canonical-scene units); $\mathrm{smooth} = \exp\bigl(-\,\mathrm{mean}_{\mathcal{D}}^{\mathrm{conf}}\|v_{t+1}-v_t\| / (m+\epsilon)\bigr)$ (Eq.~\ref{eq:smooth}); and $\mathrm{rigid} = \exp\bigl(-\tfrac{k_{\text{rough}}}{2}(\mathrm{mean}_{\mathcal{D}_x}\|\nabla_x v_t\| + \mathrm{mean}_{\mathcal{D}_y}\|\nabla_y v_t\|)\bigr)$ with $k_{\text{rough}}\!=\!400$ (Eq.~\ref{eq:rigid}). The scene-flow magnitude $m$ is the within-clip confidence-weighted mean over the top-20\% velocity mask; it is used only inside the reward at training time and is not one of the reported evaluation metrics.

\noindent\textbf{Per-axis $z$-normalization.}
Each axis $i\!\in\!\{\text{recon},\text{mot},\text{hpsv2}\}$ contributes $\tilde z[R_i] = (R_i - \mathrm{mean}_{\text{batch}}(R_i)) / \mathrm{std}_{\text{batch}}(R_i)$ to the GRPO advantage, independently per minibatch. Per-axis $z$-norm decouples axis scales: e.g.\ HPSv2 sits at ${\sim}0.20$ raw while the motion conjunction sits at ${\sim}0.80$, but after $z$-norm both contribute on a common scale with weights $0.3$ and $1.0$.

\section{Training Details and Hyperparameters}
\label{sec:supp_hyper}

Table~\ref{tab:hyperparameters} lists the full training configuration.

\noindent\textbf{Training details.}
We fine-tune a LoRA adapter ($r{=}\alpha{=}256$) over the frozen base with AdamW ($\eta{=}10^{-5}$), mixed-precision \texttt{bf16}, 4-step distilled timesteps, group size $G{=}24$ per prompt, rolling window $L{=}21$, frame-sink $S{=}3$, and NFT trust-region $\beta{=}0.1$. Group advantages over Eq.~\ref{eq:reward} are mapped to the NFT reward $\tilde r \in [0,1]$ exactly as in Sec.~\ref{sec:method:reward} (Eq.~\ref{eq:r_tilde}). We inherit Astrolabe's stabilization recipe: an EMA-updated old policy $\theta_{\text{old}}$ for the negative reference, a selective KL penalty~\cite{he2025gardo}, and a conditional reference-policy reset on KL drift. At training time, StreamVGGT~\cite{streamvggt_2026} predicts cameras on a 26-frame linearly-subsampled version of each candidate rollout at $294\times 518$; the LPIPS-AlexNet comparison of Eq.~\ref{eq:r_recon} runs between the MoVieS re-render and the input video at $480\times 832$; HPSv2~\cite{wu2023hpsv2} runs on the generated frames directly.

\noindent\textbf{Compute.}
The reward stack is a single StreamVGGT{+}MoVieS forward per candidate rollout ($26$ frames at $294\times 518$), measured at ${\approx}7$\,s/rollout; because that one forward is shared between the reconstruction and motion axes via a per-batch cache, the motion-quality conjunction adds negligible marginal cost over a reconstruction-only reward. Each backbone trains for $150$ RL steps at effective batch $384$ ($16$ prompts $\times$ $G{=}24$ rollouts), i.e.\ ${\approx}57{,}600$ scored rollouts. End-to-end LoRA fine-tuning to the reported ckpt-$150$ costs ${\approx}690\,/\,770\,/\,635$ GPU-hours on Self-Forcing\,/\,Causal-Forcing\,/\,LongLive (${\approx}20$\. Rollout sampling dominates per-step wall-clock. This is a deliberately lightweight recipe: a LoRA adapter over a frozen 4-step distilled base, one shared reconstruction forward per candidate, and no per-backbone tuning beyond the three reward-axis weights.

\begin{table}[h]
\centering
\caption{Stream4D training hyperparameters. A single training-hyperparameter set is used across Self-Forcing, Causal-Forcing, and LongLive; only the three reward-axis weights vary per backbone.}
\label{tab:hyperparameters}
\resizebox{0.95\textwidth}{!}{
\begin{tabular}{llc}
\toprule
\textbf{Module} & \textbf{Hyperparameter} & \textbf{Value} \\
\midrule
\multirow{2}{*}{\textbf{Backbone}}
& Architecture & Causal Wan 2.1 (Self-Forcing / Causal-Forcing / LongLive) \\
& Video resolution ($H \times W$) & $480 \times 832$ \\
\midrule
\multirow{3}{*}{\textbf{LoRA fine-tuning}}
& Rank $r$ / scaling $\alpha$ & 256 / 256 \\
& Dropout & 0.0 \\
& Gradient checkpointing & True \\
\midrule
\multirow{6}{*}{\textbf{Optimization}}
& Hardware & 4 nodes $\times$ 8$\times$H200, DDP \\
& Precision & bf16 \\
& Optimizer & AdamW ($\beta_1{=}0.9, \beta_2{=}0.999$) \\
& Learning rate / weight decay & $10^{-5}$ / $10^{-4}$ \\
& Max gradient norm & 1.0 \\
& Distillation timesteps $T$ & 4, sampled from $\{1000, 750, 500, 250\}$ \\
\midrule
\multirow{3}{*}{\textbf{NFT-GRPO}}
& Trust-region $\beta$ & 0.1 \\
& Per-axis $z$-normalization & enabled \\
& Dynamic-phase schedule & disabled \\
\midrule
\multirow{3}{*}{\textbf{Rollout}}
& Group size $G$ per prompt & 24 \\
& Rolling window $L$ / frame-sink $S$ & 21 / 3 \\
& Window selection & Random \\
\midrule
\multirow{4}{*}{\textbf{Reward backbone}}
& Camera estimator & StreamVGGT~\cite{streamvggt_2026} \\
& 4D Gaussian-Splatting reconstructor & MoVieS~\cite{lin2025movies} (26-frame subsample, $294\!\times\!518$) \\
& Recon comparison & LPIPS-AlexNet between re-render and input \\
& Perceptual anchor & HPSv2~\cite{wu2023hpsv2} on input rollout frames \\
\midrule
\multirow{4}{*}{\textbf{Reward stack}}
& Axes & $\{R_{\text{recon}}, R_{\text{mot}}, R_{\text{hpsv2}}\}$ (Eq.~\ref{eq:reward}) \\
& Per-axis weights $(w_{\text{recon}}, w_{\text{mot}}, w_{\text{hpsv2}})$ & SF $(1.0,\,1.0,\,0.3)$; CF $(1.0,\,0.5,\,0.6)$; LL $(0.8,\,1.0,\,0.6)$ \\
& Gauss-gate $(m_{\text{nat}}, \sigma)$ & $(0.020,\,0.010)$ \\
& Rigidity constant $k_{\text{rough}}$ & $400$ \\
\bottomrule
\end{tabular}
}
\end{table}

\section{Algorithm}
\label{sec:algorithm}
Algorithm~\ref{alg:streaming_diffusionnft} gives the full Stream4D training loop with the deployed three-axis reward (Eq.~\ref{eq:reward}). The reward backbone is a single MoVieS forward per candidate rollout, shared between the reconstruction and motion-quality axes via a per-batch reward cache.

\begin{algorithm}[t]
  \caption{Stream4D: Streaming Forward-Process RL with a 4D-Consistency Reward}
  \label{alg:streaming_diffusionnft}
  \begin{algorithmic}[1]
  \Require Distilled AR policy $\pi_\theta$, behavior $\pi_{\theta_{\text{old}}}$, KL reference $\pi_{\theta_{\text{ref}}}$;
            camera estimator (StreamVGGT); 4D-GS reconstructor (MoVieS);
            prompts $\mathcal{D}$; rolling window $L$, frame sink $S$, total windows $N$;
            reward weights $(w_{\text{recon}}, w_{\text{mot}}, w_{\text{hpsv2}})$
  \Ensure Optimized policy $\pi_\theta$
  \State Initialize $\theta_{\text{old}} \leftarrow \theta$, $\theta_{\text{ref}} \leftarrow \theta$, risk buffer $\mathcal{B} \leftarrow \emptyset$, $\rho \leftarrow \rho_0$
  \For{each epoch $k$}
      \State Sample prompts $\{c_i\}_{i=1}^B \sim \mathcal{D}$
      \State \textbf{// Phase 1: Full $N$-window streaming rollout, $G$ candidates each}
      \For{$n = 1$ to $N$}
          \State Form context $\mathcal{C}_n$ from $S$-frame sink and $L$ rolling frames
          \State Decode $G$ candidates $\{W_n^{(i,j)}\}_{j=1}^G \sim \pi_{\theta_{\text{old}}}(\cdot \mid \mathcal{C}_n, c_i)$ (shared prefix)
          \State Update KV cache with chosen $W_n$ (one $j$ per rollout)
      \EndFor
      \State \textbf{// Phase 2: 4D-consistency reward over the rollout}
      \For{each rollout $(i,j)$}
          \State $C_{1:T} \leftarrow \mathrm{StreamVGGT}(W_{1:N}^{(i,j)})$ \Comment{per-frame cameras, 26-frame subsample}
          \State $(G_{\text{can}}, A_{1:T}, \Delta_{1:T}, P, \mathrm{conf}) \leftarrow \mathrm{MoVieS}(W_{1:N}^{(i,j)}, C_{1:T})$ \Comment{canonical 4D-GS + scene flow}
          \State Re-render $\tilde W_{1:T}^{(i,j)}$ from $G_{\text{can}}\!\oplus\!A_t\!\oplus\!\Delta_t$ at $C_t$
          \State $R_{\text{recon}}^{(i,j)} \leftarrow$ LPIPS score on $(\tilde W, W)$ \Comment{Eq.~\ref{eq:r_recon}}
          \State $R_{\text{mot}}^{(i,j)} \leftarrow g(m)\cdot\mathrm{smooth}\cdot\mathrm{rigid}$ from $(P, \mathrm{conf})$ \Comment{Eq.~\ref{eq:r_mot}}
          \State $R^{(i,j)} \leftarrow w_{\text{recon}}\tilde z[R_{\text{recon}}] + w_{\text{mot}}\tilde z[R_{\text{mot}}] + w_{\text{hpsv2}}\tilde z[R_{\text{hpsv2}}]$ \Comment{Eq.~\ref{eq:reward}}
      \EndFor
      \State Compute group advantages $A^{(i,j)} \leftarrow R^{(i,j)} - \tfrac{1}{G}\sum_h R^{(i,h)}$
      \State \textbf{// Reward uncertainty via rank disagreement (selective KL)}
      \State $\mathcal{M}^{(i,j)} \leftarrow$ high-uncertainty mask from rank disagreement~\cite{he2025gardo}
      \State \textbf{// Phase 3: Window-local NFT update with rollout-level advantage}
      \For{each training window $n \in \{1,\dots,N\}$}
          \State Re-roll forward to step $n$; detach KV cache of $W_{<n}$ as constant
          \For{each mini-batch over $(W_n^{(i,j)}, A^{(i,j)}, \mathcal{M}^{(i,j)})$}
              \State Sample $t \sim \mathcal{U}(\mathcal{T}_{\text{distill}})$, $\;W_n^t \leftarrow (1-t)W_n + t\epsilon$, $\;v_{\text{target}} \leftarrow \epsilon - W_n$
              \State $v_\theta, v_{\theta_{\text{old}}}, v_{\theta_{\text{ref}}} \leftarrow \pi_{\cdot}(W_n^t, t, \mathcal{C}_n, c)$
              \State $v^+ \leftarrow \beta v_\theta + (1-\beta) v_{\theta_{\text{old}}}$, $\;v^- \leftarrow (1+\beta) v_{\theta_{\text{old}}} - \beta v_\theta$
              \State $\tilde{r}^{(i,j)} \leftarrow \mathrm{clip}(A^{(i,j)}/A_{\max},\,-1,\,1)/2 + 1/2$
              \State $\mathcal{L}_{\text{policy}} \leftarrow \tilde{r}\,\|v^+ - v_{\text{target}}\|^2 + (1-\tilde{r})\,\|v^- - v_{\text{target}}\|^2$
              \State $\mathcal{L}_{\text{KL}} \leftarrow \tfrac{1}{|\mathcal{M}|}\sum_{(i,j):\mathcal{M}^{(i,j)}=1} \|v_\theta^{(i,j)} - v_{\theta_{\text{ref}}}^{(i,j)}\|^2$
              \State $\theta \leftarrow \theta - \eta\,\nabla_\theta(\mathcal{L}_{\text{policy}} + \lambda_{\text{KL}}\mathcal{L}_{\text{KL}})$
          \EndFor
      \EndFor
      \If{$\mathcal{L}_{\text{KL}} > \tau_{\text{KL}}$ \textbf{or} $k - k_{\text{last}} > K_{\max}$}
          \State $\theta_{\text{ref}} \leftarrow \theta$, $\;k_{\text{last}} \leftarrow k$
      \EndIf
      \State $\theta_{\text{old}} \leftarrow \gamma\,\theta_{\text{old}} + (1-\gamma)\,\theta$
  \EndFor
  \end{algorithmic}
\end{algorithm}

\section{Evaluation Protocol Details}
\label{sec:supp_eval}

\noindent\textbf{Evaluation sets.}
The main-paper results (Tables~\ref{tab:compare_short},~\ref{tab:ablation_axes}, and~\ref{tab:ablation_motion}) use a 500-prompt motion-prominent subset of VidProM~\cite{wang2024vidprom}, disjoint from the training prompts. We choose VidProM rather than the World-R1 evaluation set because VidProM's prompts include substantial moving-subject content (people walking, vehicles, non-rigid foliage) where our reward is designed to operate; the World-R1 set is dominated by camera-tour-through-static-scenes prompts that bias toward rigid-scene methods.

\noindent\textbf{Model versions.}
All learned components are public releases used unmodified, with identical weights at training and evaluation time: MoVieS~\cite{lin2025movies} official release checkpoint (\texttt{movies\_ckpt.safetensors}); StreamVGGT~\cite{streamvggt_2026} public checkpoint (HuggingFace \texttt{lch01/StreamVGGT}); 4DGT~\cite{xu20254dgt} released full model (\texttt{4dgt\_full.pth}, level-of-detail config \texttt{tlod-l3}); VideoReward~\cite{liu2025improvingvideogenerationhuman} released Qwen2-VL-7B checkpoint; HPSv2~\cite{wu2023hpsv2} checkpoint \texttt{HPS\_v2.1\_compressed.pt} (v2.1). The LLM judge's model string and run dates are given in Appendix~\ref{sec:supp_judge}.

\noindent\textbf{4D-PSNR / SSIM / LPIPS.}
We re-run MoVieS 4D Gaussian-Splatting on each generated rollout at evaluation time, using StreamVGGT for camera estimation followed by MoVieS reconstruction over 26 frames at a resolution of $294 \times 518$. We then re-render the reconstructed 4D scene from the estimated camera corresponding to each input frame and compare the resulting rendering $\hat{\mathbf{I}}_t$ against the original generated frame $\mathbf{I}_t$. We report standard image-space PSNR, SSIM, and LPIPS-AlexNet for each frame and average the resulting scores over the full rollout.

For PSNR, we first compute the mean-squared error over all RGB pixels:
\begin{equation}
    \mathrm{MSE}_t
    =
    \frac{1}{3HW}
    \left\|
        \mathbf{I}_t - \hat{\mathbf{I}}_t
    \right\|_2^2.
\end{equation}
Assuming RGB values are normalized to $[0,1]$, the per-frame PSNR is
\begin{equation}
    \mathrm{PSNR}_t
    =
    10 \log_{10}
    \left(
        \frac{1}{\mathrm{MSE}_t}
    \right).
\end{equation}
We report 4D-PSNR by averaging the per-frame PSNR values:
\begin{equation}
    \mathrm{4D\text{-}PSNR}
    =
    \frac{1}{T}
    \sum_{t=1}^{T}
    \mathrm{PSNR}_t.
\end{equation}
Thus, ``4D-PSNR'' does not denote a new PSNR formulation; rather, it refers to standard PSNR computed after fitting and re-rendering the rollout with the dynamic 4D Gaussian-Splat representation.

SSIM measures local structural similarity between $\mathbf{I}_t$ and $\hat{\mathbf{I}}_t$. For corresponding local image windows $x$ and $y$, SSIM is computed as
\begin{equation}
    \mathrm{SSIM}(x,y)
    =
    \frac{
        (2\mu_x\mu_y + C_1)
        (2\sigma_{xy} + C_2)
    }{
        (\mu_x^2 + \mu_y^2 + C_1)
        (\sigma_x^2 + \sigma_y^2 + C_2)
    },
\end{equation}
where $\mu_x$ and $\mu_y$ denote the local means, $\sigma_x^2$ and $\sigma_y^2$ denote the local variances, $\sigma_{xy}$ denotes the local covariance, and $C_1$ and $C_2$ are numerical-stability constants. We spatially average the local SSIM values to obtain a per-frame score and then average across frames:
\begin{equation}
    \mathrm{4D\text{-}SSIM}
    =
    \frac{1}{T}
    \sum_{t=1}^{T}
    \mathrm{SSIM}
    \left(
        \mathbf{I}_t,
        \hat{\mathbf{I}}_t
    \right).
\end{equation}
Higher PSNR and SSIM indicate better reconstruction fidelity.

For LPIPS, we use the AlexNet-based perceptual distance. Each image pair is passed through a pretrained AlexNet feature extractor, and normalized deep features from multiple layers are compared using the learned LPIPS channel weights. Denoting the normalized feature representation at layer $l$ by $\tilde{\phi}_l(\cdot)$ and its learned channel-wise weighting by $\mathbf{w}_l$, the per-frame LPIPS distance can be written as
\begin{equation}
    \mathrm{LPIPS}_t
    =
    \sum_l
    \frac{1}{H_l W_l}
    \sum_{h,w}
    \left\|
        \mathbf{w}_l
        \odot
        \left(
            \tilde{\phi}_l(\mathbf{I}_t)_{h,w}
            -
            \tilde{\phi}_l(\hat{\mathbf{I}}_t)_{h,w}
        \right)
    \right\|_2^2,
\end{equation}
where $H_l$ and $W_l$ denote the spatial dimensions of the feature map at layer $l$. We report the average LPIPS distance across frames:
\begin{equation}
    \mathrm{4D\text{-}LPIPS}
    =
    \frac{1}{T}
    \sum_{t=1}^{T}
    \mathrm{LPIPS}_t.
\end{equation}
Unlike PSNR and SSIM, lower LPIPS indicates better perceptual agreement.

This evaluation uses the same reconstruction backbone as the training reward, so it directly measures the quantity optimized by the policy: how faithfully a generated rollout can be explained by a temporally dynamic 3D representation. We refer to it as ``4D'' reconstruction because MoVieS uses time-varying per-frame Gaussian attributes and scene flow, yielding a representation over both space and time. On the same generated rollouts, 4D reconstruction achieves approximately $5$--$7\,\mathrm{dB}$ higher PSNR than its static-3DGS counterpart, since moving content is in-distribution for the dynamic reconstructor.

\noindent\textbf{What 4DGT does and does not control for.} We call this metric \emph{reconstructor-independent}, not fully independent, and state its scope precisely. It swaps the \emph{reconstruction} model with 4DGT, so a gain that shows up under both is not an artifact of MoVieS' particular inductive biases. Yet, they both obtained cameras from the same StreamVGGT estimator.

\noindent\textbf{VideoReward.}
We use the Qwen2-VL-7B VideoReward~\cite{liu2025improvingvideogenerationhuman} model, which scores video along three heads: Visual Quality (VQ, appearance and visual fidelity), Motion Quality (MQ, motion naturalness and temporal consistency), and Text Alignment (TA, adherence to the prompt), plus an Overall score. In the main tables we report the per-prompt paired win-rate against the same backbone's distilled base: the fraction of prompts on which the method's video scores higher than the base's under each head.

\noindent\textbf{Vision-LLM judge.}
The two LLM-judge columns in the main tables come from a reward-blind vision-LLM (Gemini-3.5-Flash). \emph{Motion} is a three-way verdict on object/subject motion (preserved$=1$ / reduced$=\tfrac12$ / mostly lost$=0$), averaged over prompts to a score in $[0,1]$. \emph{Consistency win\%} is a head-to-head verdict (win$=1$ / tie$=\tfrac12$ / loss$=0$ vs.\ base) requiring both kept motion and better consistency, averaged and centered at $50\,\%$; a frozen rollout is always a loss, so consistency bought by suppressing motion earns no credit. All reported numbers are position-debiased by averaging the two presentation orders. Full judge prompts, the debiasing protocol, and reproducibility checks are given in Appendix~\ref{sec:supp_judge}.

\noindent\textbf{Baseline reproductions.}
World-R1 and VideoGPA target bidirectional T2V models, so no published checkpoints exist for the distilled-AR setting; both baselines are our reproductions on the same bases, trainer, and prompt data as our method. Reproduction configs are released with the code.

\section{LLM-Judge Details}
\label{sec:supp_judge}

This appendix documents the vision-LLM judge behind the \emph{Motion} and \emph{Consistency win\%} columns of Tables~\ref{tab:compare_short},~\ref{tab:ablation_axes}, and~\ref{tab:ablation_motion}: the exact prompts, the decoding configuration, the position-debiasing protocol, and reproducibility checks.

\noindent\textbf{Setup.}
The judge is Gemini-3.5-Flash, called via the REST API with temperature $0$ and a constrained JSON response schema with a three-way verdict enum plus a one-line free-text reason. Each call attaches two videos inline as mp4, sampled at 5\,fps. The judge never sees the reward, the scene-flow metric, the method names, or which video is ours; the prompts label the two clips only as ``BASE (reference)'' and ``EVALUATED.''

\noindent\textbf{Motion judge prompt.}
The three-way motion verdict uses the following prompt, verbatim; \texttt{\{prompt\}} is the text-to-video prompt and \texttt{\{base\_label\}} / \texttt{\{eval\_label\}} are filled with ``Video 1'' / ``Video 2'' according to the current presentation order. The verdict labels are neutral comparisons rather than loaded terms, so the judge is not primed toward a conclusion.

{\small
\begin{verbatim}
You are judging two AI-generated videos made from the SAME text prompt.

Text prompt:
"""{prompt}"""

Two videos are attached: Video 1 (first) and Video 2 (second).
- The BASE (reference) video is {base_label}.
- The video being EVALUATED is {eval_label}.

Compare the amount of MOTION in the EVALUATED video ({eval_label}) against the
BASE video ({base_label}). "Motion" means genuine object/subject motion (people
walking, cars driving, animals moving, crowds, explosions...) - NOT camera
pans/zooms over a frozen scene, and NOT flickering/texture noise.

Answer with exactly one verdict (about the EVALUATED video relative to the BASE):
  "motion_preserved"   = the EVALUATED video shows a similar amount of (or more)
                         object motion than the BASE
  "motion_reduced"     = the EVALUATED video shows noticeably less object motion
                         than the BASE, but subjects still clearly move
  "motion_mostly_lost" = the EVALUATED video is near-static or frozen while the
                         BASE clearly moves - most of the object motion is gone
Judge motion only - ignore visual quality/aesthetics differences.
\end{verbatim}
}

\noindent The reported motion score is $(\#\text{preserved} + \tfrac12\,\#\text{reduced})/N \in [0,1]$.

\noindent\textbf{Joint motion{+}consistency (Consistency) judge prompt.}
The head-to-head verdict folds motion retention and consistency into a single call; the key design choice is that a frozen evaluated video is a \emph{loss} even if it looks perfectly stable, which removes the static bias that a pure consistency rating would leave behind:

{\small
\begin{verbatim}
You are comparing two AI-generated videos made from the SAME text prompt.

Text prompt:
"""{prompt}"""

Two videos are attached: Video 1 (first) and Video 2 (second).
- The BASE model's output (the reference) is {base_label}.
- The video being EVALUATED is {eval_label}.

Decide whether the EVALUATED video ({eval_label}) is BETTER than the BASE
({base_label}). The EVALUATED video is better ONLY if it does BOTH of these,
relative to the BASE:
  (A) KEEPS THE MOTION - it preserves roughly as much genuine object/subject motion
      as the BASE (people, animals, vehicles, crowds, effects). "Motion" is real
      object/subject movement, NOT camera pans over a frozen scene and NOT texture
      flicker.
  (B) IS MORE CONSISTENT THAN THE BASE - on that moving content, objects hold their
      identity, shape, and structure BETTER than in the BASE: less morphing, warping,
      popping in/out, dissolving, or flicker than the BASE shows.

Being more consistent only counts if the motion is still there. A video that looks
cleaner only because it stopped moving has NOT kept the motion, so it is NOT better -
it is worse.

Answer with exactly one verdict (about the EVALUATED video relative to the BASE):
  "win"     = the EVALUATED video is BETTER than the BASE: it keeps about as much
              (or more) object motion AND is clearly more consistent than the BASE
              on that motion.
  "partial" = roughly a TIE: the EVALUATED video keeps the motion but is about as
              consistent as the BASE (no clear improvement either way) - neither
              clearly better nor clearly worse.
  "loss"    = the EVALUATED video is WORSE than the BASE: its motion is far behind
              the BASE (largely frozen / near-static), OR it is less consistent than
              the BASE (more morphing, identity changes, structural breakdown).
Judge motion and consistency only - ignore overall visual quality / aesthetics and
prompt wording. Consistency is a COMPARISON against the BASE, not an absolute; a
frozen or near-static EVALUATED video is always WORSE because it did not keep the
motion.
\end{verbatim}
}

\noindent The reported win-rate is $(\#\text{win} + \tfrac12\,\#\text{partial})/N$, centered at $50\,\%$; $>\!50\,\%$ means the method beats its own base.

\noindent\textbf{Position debiasing.}
Both prompts refer to the videos through the \emph{role} labels (BASE / EVALUATED) rather than the attachment slots, so the same question can be asked with the two videos in either physical order. Every pair is judged twice: once with the base attached first and once with the variant attached first, with \texttt{\{base\_label\}} / \texttt{\{eval\_label\}} swapped accordingly, so both verdicts mean ``the variant relative to the base.'' Each of the two verdicts is mapped to the three-point numeric scale (Motion: mostly-lost$=0$, reduced$=\tfrac12$, preserved$=1$; Consistency: loss$=0$, tie$=\tfrac12$, win$=1$) and the pair's reported score is the \emph{average} of its two order scores, so a video that wins in one order and ties in the other is credited $0.75$ rather than being forced to one discrete verdict. Averaging the two symmetric presentations cancels the additive position bias exactly. Reported column values are the mean of these per-pair averages over the prompt set.

\noindent\textbf{Reproducibility.}
The contested SF main-table pair (ours vs.\ World-R1) was run as three independent debiased passes; preserved\% agrees within $1$\,pt across passes (ours $61.2$--$61.4$, World-R1 $63.4$--$64.4$), and we report the final pass. Independent duplicate passes on two deployed rows differ by $\leq 0.005$ motion score and $\leq 1.2$\,pp win-rate. Per-pair verdicts with per-order records and one-line reasons are archived as JSONL alongside the aggregates.

\section{Human Study Protocol}
\label{sec:supp_human}

\noindent\textbf{Design.} A two-alternative forced-choice (2AFC) study on the LongLive backbone, which we prioritize because its $10.3$\,s horizon is where accumulated drift and the static-collapse shortcut matter most. We sample $50$ prompts from the high-motion held-out subset and form three pairings per prompt and compare Stream4D against the distilled base, the World-R1 reproduction, and VideoGPA. Each comparison is judged exactly once; the $150$ comparisons are partitioned across the $5$ raters. For each pair the rater answers two questions: (1)~which video shows \emph{more natural motion}: enough real movement, not frozen; smooth, not jittery, and (2)~which video is \emph{better overall}: jointly weighing motion amount, motion quality, and object consistency. 

\noindent\textbf{Quality control.} Each rater additionally sees two interleaved attention-check trials with an obviously frozen clip vs.\ a moving one; a rater is excluded if either check is failed. All $5$ raters passed both checks, so all $5$ are retained and none are excluded; the reported numbers therefore use the full $50$ judgments per comparison.

\section{Random-Subset Robustness Check}
\label{sec:supp_rand500}

The main evaluation (Table~\ref{tab:compare_short}) uses a motion-prominent filtered subset, chosen because it is the regime where the static-collapse shortcut matters. Table~\ref{tab:compare_rand500} repeats the comparison on a uniformly random 500-prompt subset of Vidprom, using the identical protocol. Because the pool is dominated by lower-motion prompts, the frozen VideoGPA and World-R1 sweep reconstruction and even VideoReward, yet under the motion-aware judge they lose to their own bases. Stream4D is the only method that beats its base under the joint verdict on all three backbones, and its judge scores are nearly unchanged from the motion-prominent subset (SF motion $0.816$ vs $0.833$; LL Consistency $68.5$ vs $74.2$).

\begin{table*}[!t]
\small
\centering
\caption{\textbf{Robustness check: main comparison on a uniformly random 500-prompt subset} of the same held-out VidProM set (seed-0 sample, no motion filter; $122$ prompts overlap the high-motion subset of Table~\ref{tab:compare_short}). Metrics and conventions as in Table~\ref{tab:compare_short} (\emph{4DGT}: reconstructor-independent dynamic-reconstruction cross-check, Appendix~\ref{sec:supp_eval}); all RL'd rows at ckpt-150, judged against the same backbone's base. $^{\ddagger}$The VideoGPA row uses the standard VideoGPA recipe (the best-weight variant of Table~\ref{tab:compare_short} was generated only on the high-motion subset); on this unfiltered, low-motion-dominated subset its clean freeze sweeps \emph{both} reconstructors. $^{\dagger}$marks methods whose Consistency win-rate falls below the $50\,\%$ break-even against their own base: on this low-motion-dominated subset, their reconstruction and VideoReward advantages are bought by suppressing motion (motion scores $0.14$--$0.48$), not by generating better video. Bold = best RL'd row per backbone block; shaded = ours.}
\label{tab:compare_rand500}
\setlength{\tabcolsep}{3pt}
\renewcommand{\arraystretch}{1.07}
\resizebox{\textwidth}{!}{%
\begin{tabular}{l|ccc|ccc|cc|cccc}
\toprule
\multirow{2}{*}{\textbf{Method}}
 & \multicolumn{3}{c|}{\textbf{MoVieS Recon.}}
 & \multicolumn{3}{c|}{\textbf{4DGT Recon.}}
 & \multicolumn{2}{c|}{\textbf{LLM judge}}
 & \multicolumn{4}{c}{\textbf{VideoReward win\% }} \\
\cmidrule(lr){2-4}\cmidrule(lr){5-7}\cmidrule(lr){8-9}\cmidrule(lr){10-13}
 & \textbf{PSNR}$\uparrow$ & \textbf{SSIM}$\uparrow$ & \textbf{LPIPS}$\downarrow$
 & \textbf{PSNR}$\uparrow$ & \textbf{SSIM}$\uparrow$ & \textbf{LPIPS}$\downarrow$
 & \textbf{Motion}$\uparrow$ & \textbf{Consist.}$\uparrow$
 & \textbf{VQ}$\uparrow$ & \textbf{MQ}$\uparrow$ & \textbf{TA}$\uparrow$ & \textbf{Ovr}$\uparrow$ \\
\midrule
\multicolumn{13}{c}{\textit{Distilled AR backbone: Self-Forcing}~\cite{huang2025self} (5\,s)} \\
\midrule
Base (B0)                                                                       & 18.19 & 0.831 & 0.239 & 16.23 & 0.546 & 0.439 & ---   & ---   & ---   & ---   & ---   & --- \\
World-R1~\cite{wang2026world}                                                     & 19.35 & 0.885 & 0.182 & 16.60 & 0.560 & 0.412 & 0.773 & 68.2  & 61.4  & 49.4  & 54.2  & 56.8 \\
VideoGPA~\cite{du2026videogpa}$^{\ddagger\dagger}$                                 & \textbf{24.88} & \textbf{0.908} & \textbf{0.143} & \textbf{19.87} & 0.621 & \textbf{0.396} & 0.477 & 38.3  & \textbf{63.0}  & \textbf{63.4}  & \textbf{65.0}  & \textbf{67.4} \\
\rowcolor{ourscolor} $+$ \textbf{Stream4D (Ours)}                              & 21.23 & 0.895 & 0.177 & 17.70 & \textbf{0.657} & 0.411 & \textbf{0.816} & \textbf{75.1} & 59.8  & 56.6  & 54.4  & 59.8 \\
\midrule
\multicolumn{13}{c}{\textit{Distilled AR backbone: Causal-Forcing}~\cite{zhu2026causal} (5\,s)} \\
\midrule
Base (B0)                                                                       & 16.30 & 0.802 & 0.252 & 14.94 & 0.512 & 0.455 & ---   & ---   & ---   & ---   & ---   & --- \\
World-R1~\cite{wang2026world}                                                     & 19.67 & \textbf{0.894} & 0.162 & 16.34 & 0.575 & 0.417 & 0.637 & 61.1  & \textbf{72.6}  & 60.6  & \textbf{57.0}  & 65.8 \\
VideoGPA~\cite{du2026videogpa}$^{\ddagger\dagger}$                                 & \textbf{22.94} & 0.843 & \textbf{0.147} & \textbf{19.33} & 0.605 & \textbf{0.377} & 0.484 & 36.5  & 67.6  & 54.8  & 53.4  & 62.2 \\
\rowcolor{ourscolor} $+$ \textbf{Stream4D (Ours)}                              & 20.99 & 0.890 & 0.156 & 17.49 & \textbf{0.650} & 0.404 & \textbf{0.716} & \textbf{64.9} & 71.4  & \textbf{65.8}  & 52.8  & \textbf{68.6} \\
\midrule
\multicolumn{13}{c}{\textit{Distilled AR backbone: LongLive}~\cite{yang2025longlive} (10.3\,s)} \\
\midrule
Base (B0)                                                                       & 18.11 & 0.863 & 0.217 & 15.45 & 0.509 & 0.447 & ---   & ---   & ---   & ---   & ---   & --- \\
World-R1~\cite{wang2026world}$^{\dagger}$                                          & 23.01 & 0.940 & 0.137 & 17.53 & 0.630 & 0.438 & 0.457 & 46.8  & \textbf{74.8}  & 68.6  & 63.4  & 75.8 \\
VideoGPA~\cite{du2026videogpa}$^{\ddagger\dagger}$                                 & 22.99 & \textbf{0.943} & \textbf{0.130} & 19.24 & \textbf{0.668} & \textbf{0.339} & 0.136 & \phantom{0}9.8 & 66.0  & 63.6  & 54.6  & 67.0 \\
\rowcolor{ourscolor} $+$ \textbf{Stream4D (Ours)}                              & \textbf{24.60} & 0.910 & 0.147 & \textbf{19.59} & 0.646 & 0.417 & \textbf{0.701} & \textbf{68.5} & \textbf{74.8}  & \textbf{76.6}  & \textbf{65.4}  & \textbf{78.2} \\
\bottomrule
\end{tabular}
}
\raggedright\footnotesize\textit{On this unfiltered subset the frozen specialists sweep the reconstruction metrics under \emph{both} reconstructors---SF VideoGPA$^{\ddagger}$ takes the best MoVieS \emph{and} 4DGT PSNR (and the best VideoReward-Overall)---because a clean freeze of a near-static scene reconstructs well under any reconstructor. The judge is what exposes the shortcut: VideoGPA's motion score is $0.477$ and it loses to its own base on Consistency ($38.3\,\%$), as do CF VideoGPA$^{\ddagger}$ ($36.5\,\%$), LL VideoGPA$^{\ddagger}$ ($9.8\,\%$), and LL World-R1 ($46.8\,\%$). Stream4D is the only method that beats its base under the joint verdict on all three backbones. Joint verdicts cover $499/500$ prompts per row (one API failure each).}
\vspace{-2mm}
\end{table*}

\section{Additional Qualitative Results}
\label{sec:supp_qual}

\begin{figure*}[t]
\centering
\includegraphics[width=\textwidth]{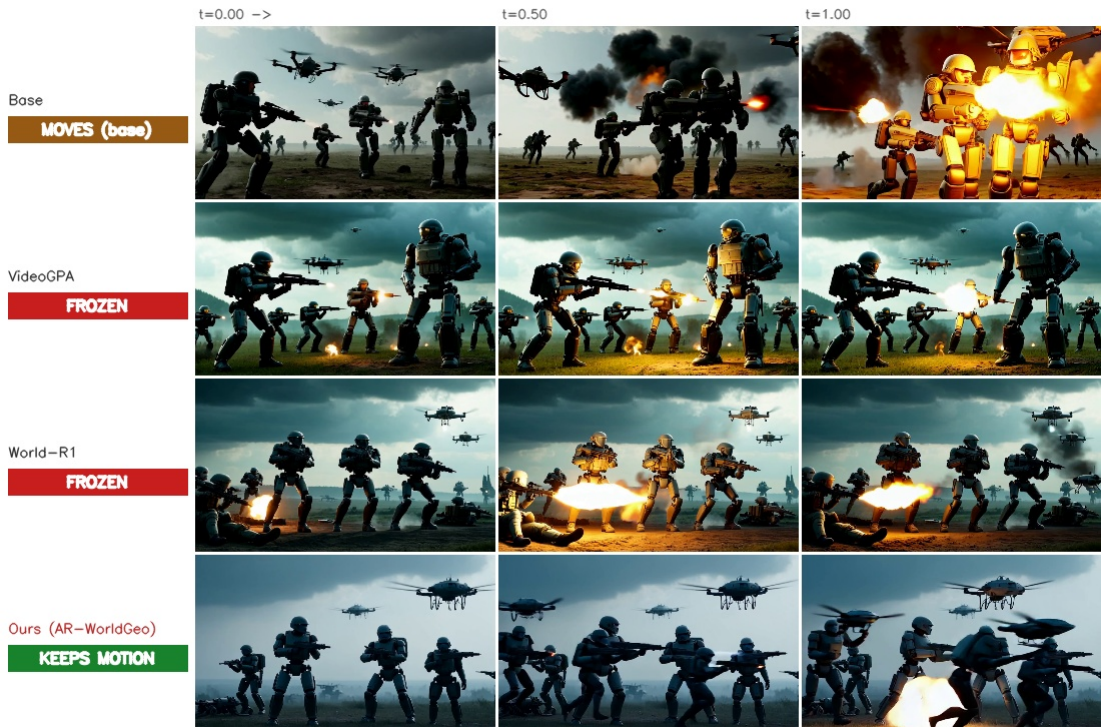}\\[3pt]
\includegraphics[width=\textwidth]{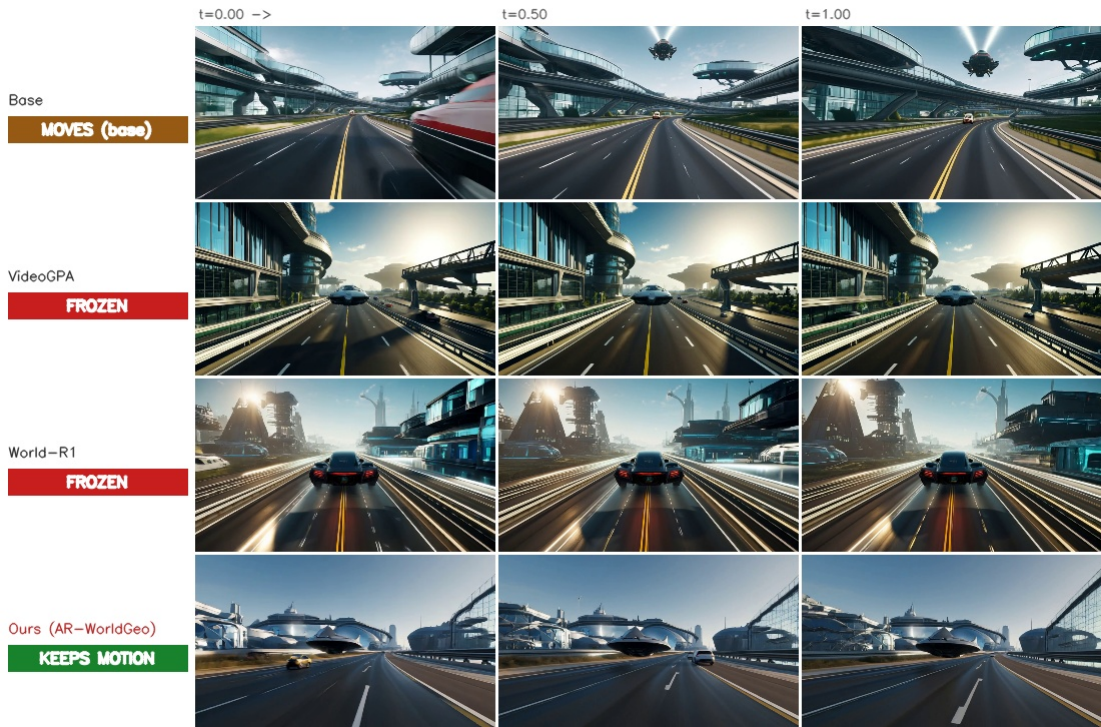}\\[3pt]
\includegraphics[width=\textwidth]{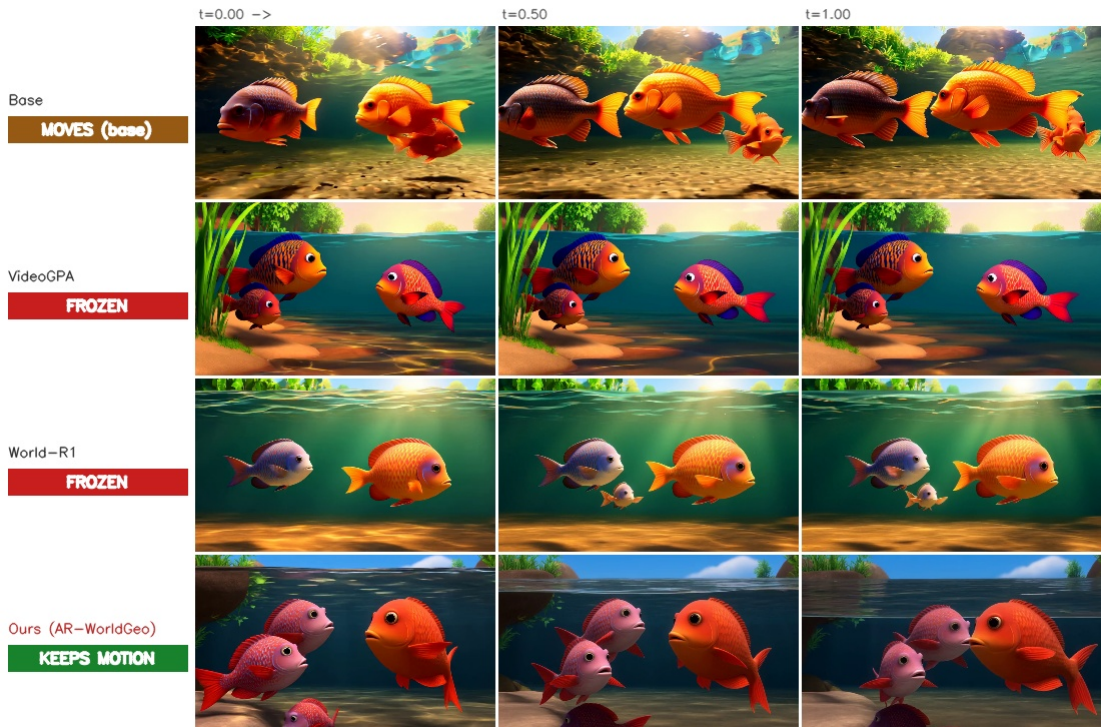}
\caption{\textbf{Motion preserved vs.\ frozen: the static-reward shortcut, one prompt per backbone.} Top: Causal-Forcing (a WW3 battle); middle: Self-Forcing (a futuristic highway); bottom: LongLive (fish in a river). Three uniformly-spaced frames per method, all RL'd rows at ckpt-150; read each row \emph{across} time. The static-3DGS baselines (VideoGPA, World-R1) are frozen---subjects hold the same pose from $t{=}0$ to $t{=}1$---while the distilled bases move but deform, and Stream4D keeps the motion while holding object identity and scene structure. Table~\ref{tab:compare_short} quantifies motion preservation over the full prompt set.}
\label{fig:qualitative}
\end{figure*}

\begin{figure*}[t]
\centering
\includegraphics[width=\textwidth]{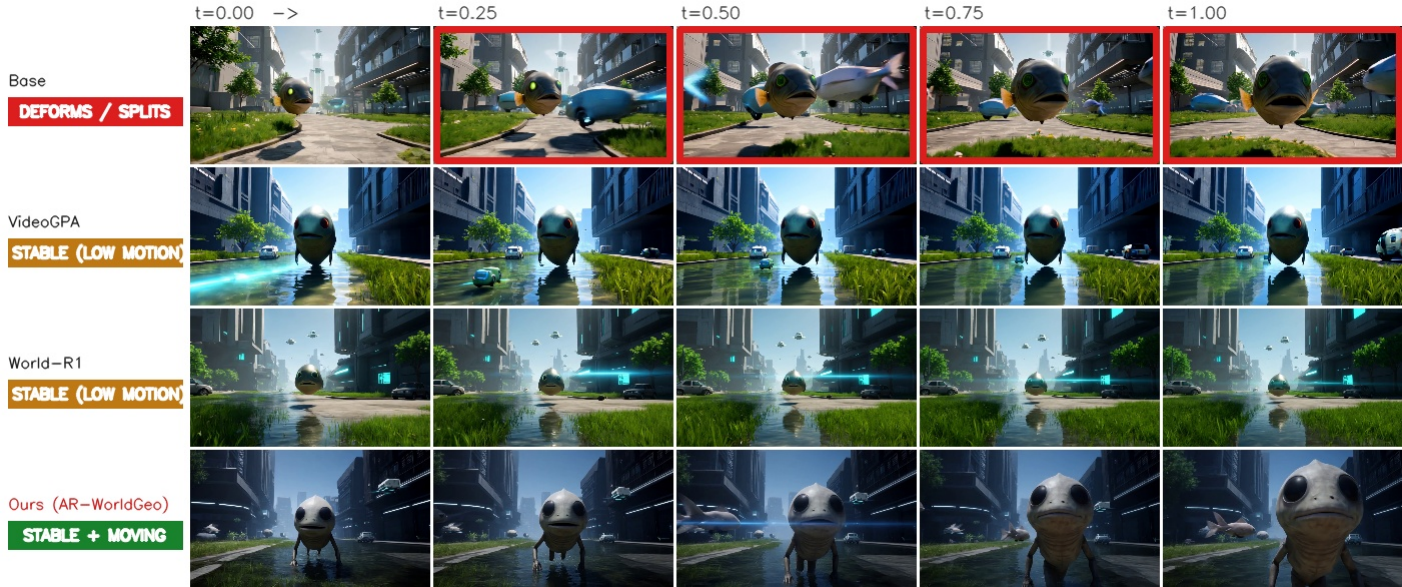}
\caption{\textbf{Object-identity under motion: the base breaks the object apart, the static-reward baselines freeze the scene, and only ours keeps a \emph{moving} object intact.}}
\label{fig:consistency}
\end{figure*}

\begin{figure*}[t]
\centering
\includegraphics[width=\textwidth]{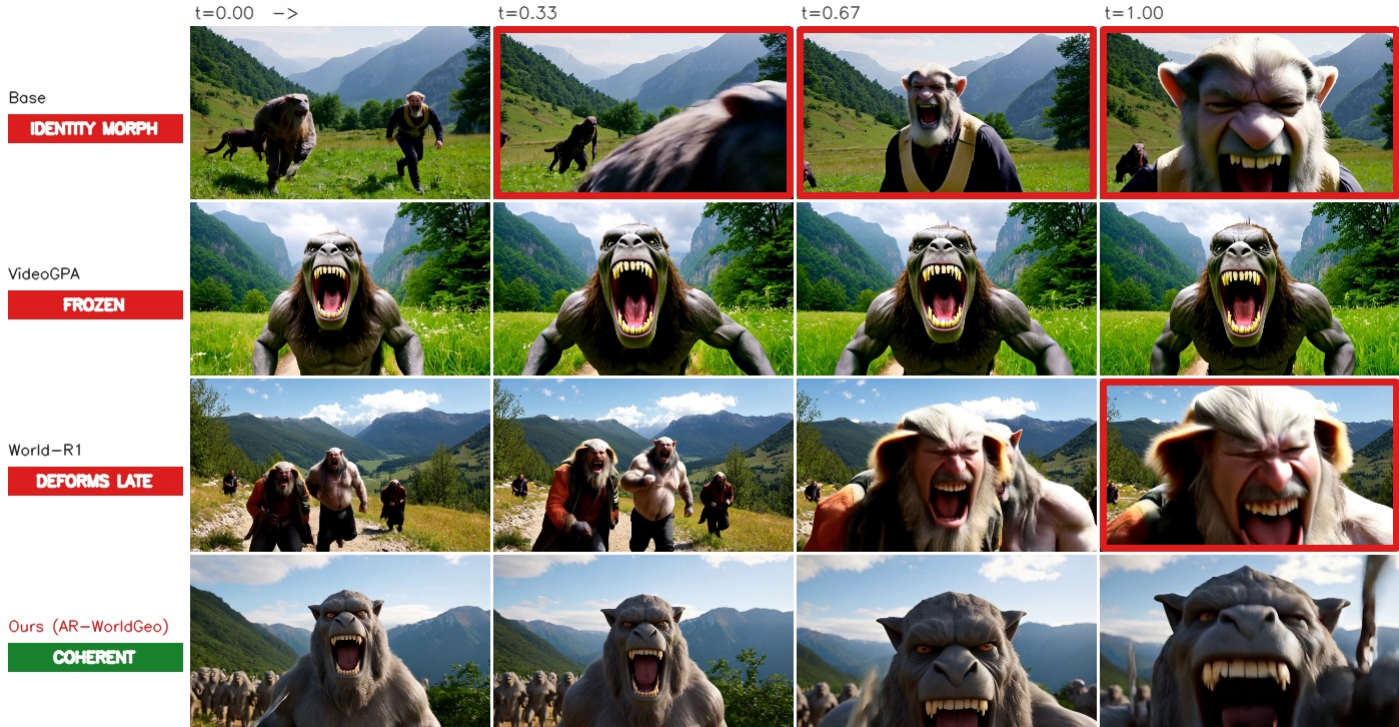}
\caption{\textbf{Consistency companion on Self-Forcing: an ancient-battlefield beast} Red outlines mark failure frames and the coloured badge names each row's behaviour. The failures are complementary: the \emph{base} moves but \textbf{morphs}; the static-reward baselines are either \textbf{frozen}. Stream4D stays \textbf{coherent}: one beast, one identity, approaching the camera.}
\label{fig:consistency_supp2}
\end{figure*}

\begin{figure*}[t]
\centering
\includegraphics[width=\textwidth]{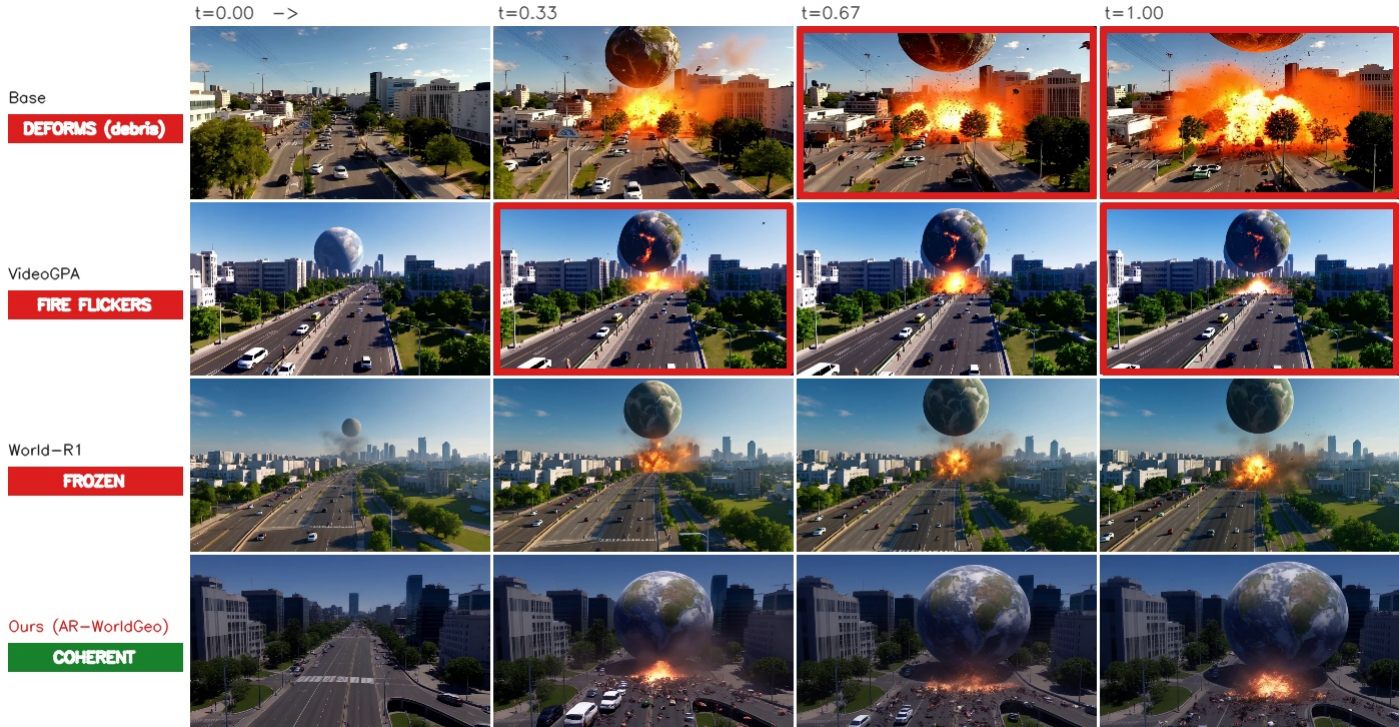}
\caption{\textbf{Consistency companion on LongLive: a planet descending over a city} The base \textbf{deforms} into spreading debris/fire; VideoGPA's explosion \textbf{flickers} in and out over an otherwise static scene; World-R1 is \textbf{frozen}. Stream4D keeps a \textbf{coherent} descending planet over a stable city.}
\label{fig:consistency_supp}
\end{figure*}

Figure~\ref{fig:qualitative} shows motion-preservation filmstrips for one prompt per backbone (uniformly-spaced frames). Figures~\ref{fig:consistency}--\ref{fig:consistency_supp} show the complementary failure, identical timestamps for every method: rollouts that move but deform, rollouts that buy stability by freezing, and Stream4D avoiding both.

\section{Limitations}
\label{sec:exp:limits}

(i)~The judge metrics come from a single vision-LLM; they are corroborated by VideoReward, the 4D metrics. (ii)~4DGT score rules out dependence on MoVieS' training bias, but they still share the VGGT component. 

\end{document}